Article

# Unsupervised crack detection on complex stone masonry surfaces

*Panagiotis Agrafiotis*, Anastastios Doulamis, Andreas Georgopoulos*

*National Technical University of Athens, School of Rural and Surveying Engineering, Lab. of Photogrammetry Zografou Campus, 9 Heroon Polytechniou str., 15780, Zografou, Athens, Greece*

A B S T R A C T

Computer vision for detecting building pathologies has interested researchers for quite some time. Vision-based crack detection is a non-destructive assessment technique, which can be useful especially for Cultural Heritage (CH) where strict regulations apply and, even simple, interventions are not permitted. Recently, shallow and deep machine learning architectures applied on various types of imagery are gaining ground. In this article a crack detection methodology for stone masonry walls is presented. In the proposed approach, crack detection is approached as an unsupervised anomaly detection problem on RGB (Red Green Blue) image patches. Towards this direction, some of the most popular state of the art CNN (Convolutional Neural Network) architectures are deployed and modified to binary classify the images or image patches by predicting a specific class for the tested imagery; "Crack" or "No crack", and detect and localize those cracks on the RGB imagery with high accuracy. Testing of the model was performed on various test sites and random images retrieved from the internet and collected by the authors and results suggested the high performance of specific networks compared to the rest, considering also the small numbers of epochs required for training. Those results met the accuracy delivered by more complex and computationally heavy approaches, requiring a large amount of data for training. Source code is available on GitHub: https://github.com/pagraf/Crack-detection while datasets are available on Zenodo: https://doi.org/10.5281/zenodo.6516913.

## 1 Introduction and aims

Masonry structures represent the highest proportion of building stock worldwide, including Cultural Heritage assets and historical buildings. Currently, the structural condition of such structures and especially of monuments, is predominantly manually inspected which is a laborious, costly and subjective process [1]. With recent developments in Computer Vision and Machine Learning, there is a great opportunity to exploit RGB and/or Multi-Spectral imagery to speed up this process, with increased objectivity, high precision and accuracy. Computer Vision for detecting building pathologies has interested researchers for quite some time. Vision-based crack detection is a non-destructive assessment technique, which can be useful especially for Cultural Heritage where strict regulations apply and even simple interventions, such as placing crack-rulers, are not permitted by the conservation authorities [1]. Lately, shallow and deep Machine Learning architectures applied on various types of imagery (RGB, RGB-Depth, multi-spectral etc.) are gaining ground due to the increased automation and accuracy they offer.

CNNs are recently given great attention because of their extended applications in image classification, semantic segmentation, and other fundamental computer vision problems. They usually consist of the feature extraction part, which is made of convolutional layers and pooling layers, and the classification part, containing many stacked fully connected layers. In the first part, kernels in convolutional layers manipulate the input image, multiplying the weights in each kernel by the pixels' values and combining the sum to create a new image-like array passed to the next layer. Pooling layers play a role in down-sampling to reduce the number of data and save computational resources. In the next part, the image first passes through a flatten layer to be converted to a one-dimensional array. The following fully connected layers use this array as input and produce the predicted label by applying the linear combination and the non-linear activation function. Because of their advantage of extracting deep features layer by layer, nowadays, CNNs are widely used to solve real-world problems.

CNN based image classification can be categorized into three types: (I) image or image patch classification (Figure 1a, c), (II) boundary box regression (Figure 1b) and (III) semantic segmentation (Figure 1d) [3]. In image classification, the image or image patch is labelled with a class. When boundary box regression is considered, a box bounds the detected object, which in the discussed cases is a crack, and reveals its position and boundaries. To achieve this, the weights of the last dense layer are exploited. These two classification techniques have been extensively used to detect cracks and other defects delivering very promising results [4-7]. These techniques are implemented at block/patch level rather than at pixel level. A combination of the above two classification types is performed for crack detection in the presented research, since a class for the image or the image patch is predicted, also delivering the position of the cracks by projecting the weights of the last dense layer in the form of an attention map.

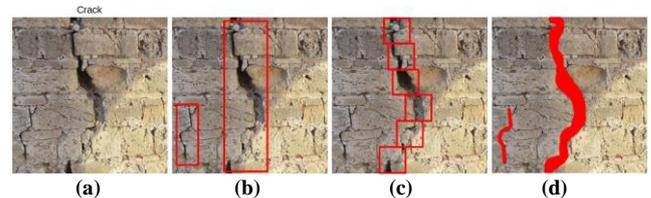

**Fig. 1 - Crack detection with image classification (a), with boundary box regression (b), with image patch classification (c) and with pixel semantic segmentation (d)**

The exact location can also be determined in high accuracy using semantic segmentation methods. They can deliver the width or length of any defects/cracks since each pixel is assigned to a class label [8-11]. Fully Convolutional Networks (FCNs), have been extensively used for semantic segmentation in many applications [15]. FCNs performed as an extended CNN, where the final prediction was a semantically segmented image instead of a class identification [1]. Recently, FCNs have been used widely for semantic segmentation on images containing cracks [16-19, 37, 38]. Feature Pyramid Networks (FPNs) are typical model architectures to generate pyramidal feature representations for object detection. These architectures aim at extracting various features at different scales and then fuse them leading to pixel-level class predictions of higher accuracy [20]. FPNs have also been used widely for crack detection [21, 22]. However, training those networks requires a large amount of manually annotated data, a costly and time-consuming procedure.

To deal with the above issue, transfer learning has been extensively implemented on different fields of computer vision with remarkable results and is considered suitable when the training dataset is small allowing for better performance and much less computational effort. The intuition behind transfer learning for image classification is that if a model is trained on a large and general enough dataset, this model will effectively serve as a generic model of the visual world [23]. CNNs utilizing transfer learning have been used extensively for image classification and semantic segmentation of cracks [1, 10, 16, 24, and 25]. Lately, different studies obtained remarkable results in crack segmentation by implementing region proposal networks followed by algorithms for pixel-level crack detection [26, 27].

---

* Corresponding author
Email: pagraf@central.ntua.gr (P. Agrafiotis)

Despite the variability of the developed methods, they are mostly devoted to concrete, pavement, brick-walls or road crack detection problems, which are much different problems compared to crack detection on stone masonry walls of Cultural Heritage sites. In the context of HYPERION H2020 project "Development of a Decision Support System for Improved Resilience & Sustainable Reconstruction of historic areas to cope with Climate Change & Extreme Events based on Novel Sensors and Modelling Tools" (https://www.hyperion-project.eu/), an easy to deploy crack detection model is sought.

With this in mind, the presented work focuses on testing and evaluating the most popular state of the art CNN architectures and presenting a simple and easy to deploy unsupervised methodology to detect cracks on complex stone masonry surfaces where joints' visual characteristics are much similar to cracks, thus providing new paradigms for the assessment of historical structures. Towards that direction, typical transfer-learning, i.e., the model is built upon a pre-trained open-source model and training from scratch approaches are presented for each architecture and each dataset, evaluating in depth the results, both in terms of classification and crack localization, comparing also the required training time. The obtained results provide an insight for researchers working with deep learning-based algorithms for crack detection, while the developed model represents a stone masonry crack detection tool that can be scaled to more complex models, including multi-label classification. The source code of the CNN systems used is freely available on GitHub: https://link will be available upon acceptance requiring only a few data to be fine-tuned, if needed. Datasets are also available on Zenodo: https://link will be available upon acceptance

The rest of the paper is organized as follows: In Section 2, the datasets used are described, and afterwards the proposed methodology is presented in detail and justified. Section 3 discusses the tests performed and the experimental results are presented while Section 4 concludes the paper.

## 2    Dataset and Methodology

The tested CNN architectures and models and the presented methodology are intended to increase the automation level on masonry Cultural Heritage structures' inspection using RGB imagery which till now is performed manually, a laborious, costly, and subjective process. The crack detection problem, is approached as an anomaly detection problem on RGB image patches, containing cracks or not. Towards this, in this paper the performance of VGG16 [2], VGG19 [2], InceptionResNetV2 [31], MobileNetV3Small [32], MobileNetV3Large [32], DenseNet121 [33], DenseNet169 [33], DenseNet201 [33], ResNet50V2 [34], ResNet101V2 [34], and Xception [35] CNN architectures are compared and evaluated over different real world datasets, and the efficiency and adaptability in classifying images or image patches of stone masonry walls by delivering a specific class for the tested imagery; "Crack" or "No crack", and consequently detect and localize those cracks on the RGB imagery is validated.

The experimental program comprises four phases:
1    Create the reference image datasets.
2    Establish the reference CNN architectures and models.
3    Implement transfer-learning approach or
4    Implement training from scratch approach and
5    Run the training experiments and evaluate the results. The details on each phase are presented as follows.

### 2.1    Datasets

To train and evaluate the aforementioned CNN models, various images with cracks are used from the test sites of HYPERION H2020 project as well as other sites, not being CH sites and not related to the rest of the sites geographically, visually or chronologically. Specifically, square image patches of 224x224 pixels (default input size for VGG16 and VGG19 models) from the test sites of Naillac pier, St. Nikolaos as well as from the roman bridge in Rhodini (Rhodes) are used, forming the *CRACK-CH: A Crack detection and classification dataset on complex stone masonry* surfaces which is available on Zenodo: https://doi.org/10.5281/zenodo.6516913.

The Saint Nikolaos Fort is an important part of the great fortifications of the Medieval City of Rhodes located at the entrance of the Mantraki port. At this location, there was just a chapel dedicated to Saint Nikolaos until 1464, when it was turned to a fortification. Since then, it has undergone reinforcements and expansions in order to defend the city and the harbour. The outer walls were built in 1480 AD and in 1863 AD it was finally transformed to a lighthouse. The second study area is the Naillac Pier at Saint Paul's rampart where a monumental tower was located as part of the fortification of the Commercial Harbor of Rhodes. It was constructed around 1400 AD on the Hellenistic Pier, but it was destroyed in 1863 after a severe earthquake. In 2017, the Naillac Tower was graphically reconstructed and presented as it stood until 1863, during the Ottoman rule. The Rhodini Roman Bridge is one of the few ancient bridges surviving in Greece and part of the Hellenistic fortification of the city, making it a monument of great importance. It was built across the stream of Rhodini, situated outside the Medieval City and has two arched openings. The Roman Bridge is in continuous use until today and its structural integrity has deteriorated, while the scaffoldings which now support the arches are gradually rusting and losing their efficiency [30]. While Naillac and St. Nikolaos test sites formed the respective datasets (Fig. 2, Table 1), the Rhodini Roman bridge's images, due to their small number, were mixed with images crowd sourced from the internet, forming the "Random" dataset. Those images were split into two separate categories,: "Cracks" and "No cracks", facilitating the later training and evaluation of the model.

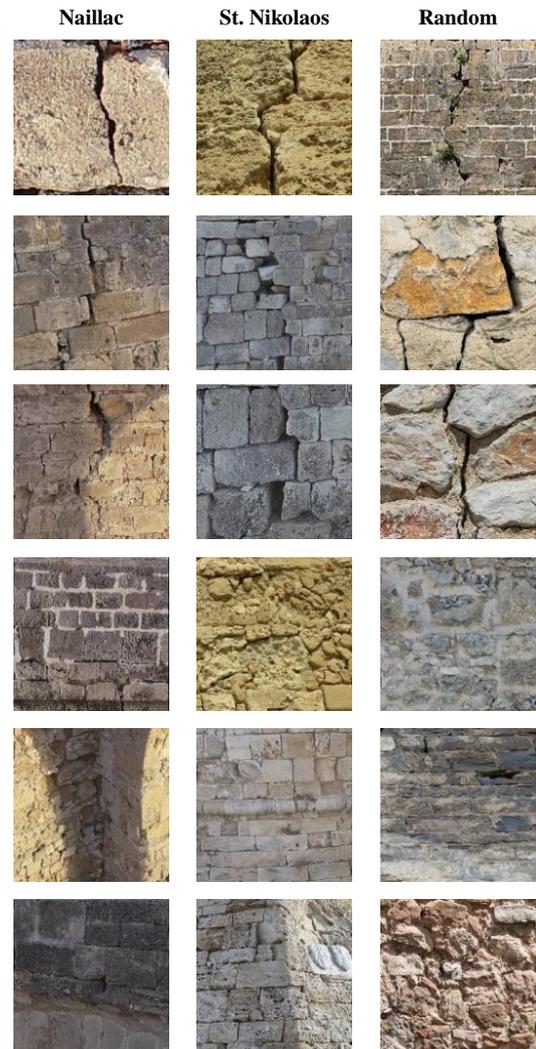

**Naillac          St. Nikolaos          Random**

**Fig. 2 – Sample images of the different test sites used.**

Sample images are shown in Fig. 2 and the number of images in each dataset in Table 1. In total, only 98 images were used. 56 of the images represent areas with various types and dimensions of cracks on various types of materials while the rest 42 represent areas without cracks, but complex enough (see Fig. 2). To augment the available data during training, vertical and horizontal flipping, colour jittering as well as random rotation was applied to the images. For VGG16 and VGG19 the images were converted from RGB to BGR, then each colour channel was zero-cantered with respect to the ImageNet dataset, without scaling. For ResNet50, ResNet101, MobileNet small and large, Xception and InceptionResNetV2 input pixel values were scaled between -1 and 1, sample-wise. Finally, for DenseNet architectures, the input pixel values were scaled between 0 and 1 and each channel is normalized with respect to the ImageNet dataset.

**Table 1: Number of images for each dataset and each label**

| Dataset | Label | # Images |
|---|---|---|
| Naillac | Crack | 22 |
| | No Crack | 14 |
| St. Nikolaos | Crack | 8 |
| | No Crack | 16 |
| Random images | Crack | 26 |
| | No Crack | 12 |
| Total images | | 98 |

### 2.2    CNN architectures and models used

Main goal of the implemented CNN models here is the classification of the images into "Crack" and "No crack" as well as the important task of detection and localization of the cracks, if any, on the imagery. To achieve this dual purpose, the most efficient method relies on a strong classifier. Initially, due to the small number of available imagery for training a full-scale model from scratch, this was decided to be achieved through a transfer learning approach. Transfer learning consists of taking features learned on one problem, and leveraging them on a new, similar problem. Next, training the models from scratch was performed to comparatively evaluate the accuracy results network-wise for both the approaches. The following deep learning models have been used. Their selection was based on accuracy and popularity criteria.

**Table 2: The implemented models, the number of their parameters in millions (M) and the depth which is the topological depth of the network. This includes activation layers, batch norm. layers etc.**

| model | Parameters | Depth |
|---|---|---|
| VGG16 [2] | 138.4M | 16 |
| VGG19 [2] | 143.7M | 19 |
| InceptionResNetV2 [31] | 55.9M | 449 |
| MobileNetV3Small [32] | 2.9M | 66 |
| MobileNetV3Large [32] | 5.4M | 217 |
| DenseNet121 [33] | 8.1M | 242 |
| DenseNet169 [33] | 14.3M | 338 |
| DenseNet201 [33] | 20.2M | 402 |
| ResNet50V2 [34] | 25.6M | 103 |
| ResNet101V2 [34] | 44.7M | 205 |
| Xception [35] | 22.9M | 81 |

VGG16 is a convolutional neural network that is 16 layers deep while VGG19 is a similar network with 19 layers depth. The network has an image input size of 224x224 pixels. Inception-ResNet-v2 is a convolutional neural architecture that builds on the Inception family of architectures but incorporates residual connections (replacing the filter concatenation stage of the Inception architecture). MobileNetV3 is a convolutional neural network that is tuned to mobile phone CPUs through a combination of hardware-aware network architecture search (NAS) complemented by the NetAdapt algorithm, and then subsequently improved through novel architecture advances. This network was tested in order to show the potential of crack detection on mobile devices. A DenseNet is a type of convolutional neural network that utilizes dense connections between layers, through Dense Blocks, where we connect all layers (with matching feature-map sizes) directly with each other. To preserve the feed-forward nature, each layer obtains additional inputs from all preceding layers and passes on its own feature-maps to all subsequent layers. Residual Networks, or ResNets, learn residual functions with reference to the layer inputs, instead of learning un-referenced functions. Residual networks let stacked layers fit a residual mapping. They stack residual blocks on top of each other to form a network: e.g. a ResNet50 has fifty layers using these blocks and a ResNet101 has one hundred and one layers using these blocks. There

is empirical evidence that these types of network are easier to optimize, and

$$precision = \frac{TP}{TP + FP}, recall = \frac{TP}{TP + FN}$$

can gain accuracy from considerably increased depth [36]. Xception is a convolutional neural network architecture that relies solely on depth-wise separable convolution layers. While standard convolution performs the channel wise and spatial-wise computation in one step, Depth-wise Separable Convolution splits the computation into two steps: depth-wise convolution applies a single convolutional filter per each input channel and point wise convolution is used to create a linear combination of the output of the depth wise convolution.

In the transfer learning approach performed, the above base models were instantiated and pre-trained weights on ImageNet [28, 29] were loaded into, to kick-start training. Then all the layers of the base models were frozen and to avoid overfitting, the top layers are being excluded from the original model and a new model was created on top of the output of the base model layers. This new model was created using a Global Average Pooling 2D layer, to keep the spatial dimension of the base models' outputs followed by a dense classifier with two units with a softmax activation function. In the training from scratch approach all the layers of the base models were unfrozen. For training the models, a Stochastic Gradient Descent (SGD) optimizer was used for VGG16 and VGG19 models while an Adam optimizer was used for the remaining models. Learning rates used in each case can be found in the Appendix. To compute the loss between the labels and predictions, the Categorical Cross Entropy loss was used. Those selections are reflecting the best results achieved for each model, after intensive testing. Since models often benefit from reducing the learning rate by a factor of 2-10 once learning stagnates, a callback that monitors the validation loss was also used. If no improvement is seen for a 'patience' number of epochs, five in our case, the learning rate is then reduced till a minimum learning rate is set.

To perform cracks localization on the test imagery, many approaches can be used. The most common one is to replace the class score by bounding box location candidates. However, in the approach presented here, since a bounding box would contain greater areas of the image instead of only the detected crack or cracks, an attention map representation is used. This way, only the detected cracks are highlighted on the imagery in "red-ish" color, providing also additional useful information. To achieve this, the weights of the final dense layer of the CNN are exploited. This activation map is then bi-linearly up-sampled to have the same size as the original RGB image, and then it is projected on it, generating the resulting images.

### 2.3    Evaluation Metrics

To evaluate the different training and testing approaches, several metrics are used; precision which gives the ability of a classification model to return only relevant instances, recall which gives the ability of a classification model to identify all relevant instances, F1 score which is a single metric that combines recall and precision using the harmonic mean and accuracy which is the ratio of the correctly labelled subjects to the whole pool of subjects. While recall expresses the ability to find all relevant instances in a dataset, precision expresses the proportion of the data points that the model says was relevant were actually relevant.

$$accuracy = \frac{TP + TN}{TP + TN + FP + FN}$$

where TP are the true positives: data points labelled as "Crack" that are actually "Cracks", FP are the false positives: data points labelled as "Cracks" that are actually "No cracks", TN are the true negatives: data points labelled as "No cracks" that are actually "No cracks" and FN are the false negatives: data points labelled as "No cracks" that are actually "Cracks". Considering crack localization, visual evaluation was performed.

## 3    Experimental Results

The 3 available datasets were used to form six different test cases. This way, the generalization potential of the networks would be highlighted while limitations and drawbacks would come into light. Table 2 presents the 6 test cases and the training and testing images used for each one. All experiments were performed on an NVIDIA GTX 1070 GPU and a 2.20GHz Intel Core i7-8750H CPU.

**Table 2: The six different test cases performed in this article.**

| Test case | Label | # Training Images | # Testing Images |
|---|---|---|---|
| 0 | Crack | 35 | 21 |
| | No Crack | 35 | 20 |
| 1 | Crack | 28 | 27 |
| | No Crack | 28 | 27 |
| 2 | Crack | 50 | 8 |
| | No Crack | 39 | 16 |
| 3 | Crack | 36 | 22 |
| | No Crack | 41 | 14 |
| 4 | Crack | 30 | 26 |
| | No Crack | 30 | 12 |
| 5 | Crack | 26 | 30 |
| | No Crack | 12 | 30 |

The same six test cases were followed both for the transfer learning and the training from scratch approach, however, below, training and validation accuracy/loss curves and typical predictions of only the latest are presented in detail, since when training from scratch, models achieved higher accuracy, especially in crack localization. Extensive details and results of all the tests performed can be found in Appendix of this article.

### 3.1 Experiments on Training - Test cases 0 and 1

The mixed dataset contains images from all the HYPERION datasets, including random images retrieved from the internet and images captured from non-CH masonry walls by the authors. 55 images of those are depicting areas of masonry walls with "No cracks" while 56 images are labelled as "Cracks", totalling 111 images. For the Test case 0, 35 images of each category were used for training the models while for the Test case 1, 28 images of each category used, forming the respective percentages of training-testing data as follows: 63% - 37% and 50% - 50% respectively. At this point it is highlighted that each image on the dataset is unique, and testing is performed always on unseen data. When following the transfer learning approach, for the Test case 0, ResNet50 with a learning rate equal to 0.000085 achieved the best accuracy 0.90, being trained for a total time of 74.141 seconds. ResNet101 and Inception-ResNet-v2 follow with 0.88 accuracy.

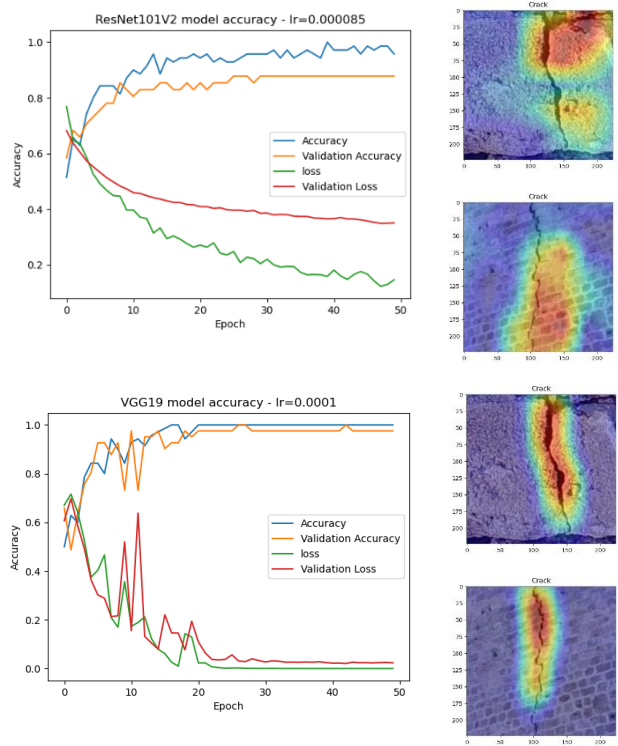

**Fig. 3 – Models' training and validation accuracy, training and validation loss and typical images from the testing datasets, demonstrating the detected crack and the predicted image class.**

Results for ResNet101 are displayed on top while results for the VGG19 are displayed on the bottom. (test case 0)

For the Test case 1, DenseNet121, DenseNet201 and ResNet50 achieved the best accuracy of 0.87, using 0.0001, 0.0001 and 0.000085 learning rates and being trained in 106.533, 197.251 and 73.580 respectively. For the training from scratch approach over Test case 0, VGG16 and VGG19 achieved the higher accuracy scores of 0.98 in 138.383 and 191.179 seconds respectively, both with a learning rate of 0.0001. For the Test case 1, only VGG19 achieved the best accuracy score, reaching 0.96, being trained for a total time of 148.327 seconds with a learning rate of 0.0001. VGG16 follows with 0.91 accuracy score and slightly less training time, 119.763 seconds.

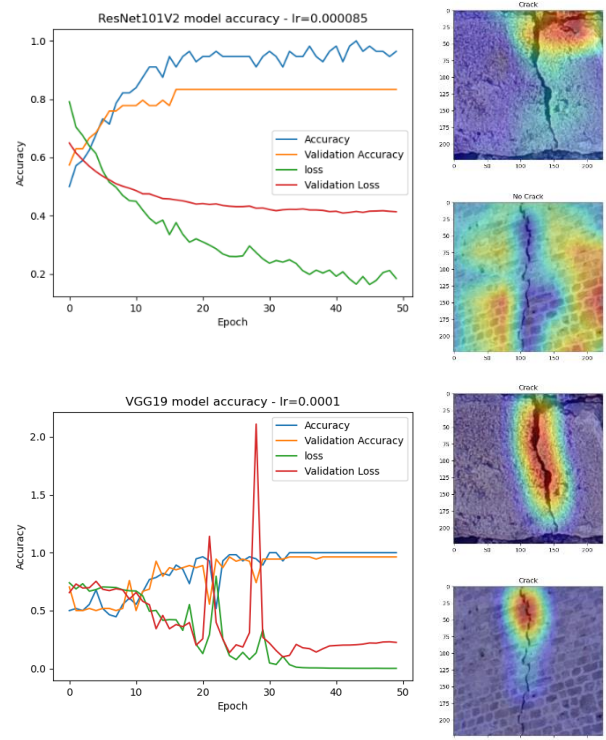

**Fig. 4 – Models' training and validation accuracy, training and validation loss and typical images from the testing datasets, demonstrating the detected crack and the predicted image class. Results for ResNet101 are displayed on top while results for the VGG19 are displayed on the bottom. (Test case 1)**

As expected, due to the small number of available training data, differences in training and validation results are apparent between test case 0 and test case 1. Even if, in the Test case 1 the maximum testing accuracy achieved reached the 0.96 and in the Test case 0, reached the 0.90, by observing Fig. 3 and Fig. 4 the effect of reducing the training data from 63% to 50% of the total data is clear. In the transfer learning approach of Test case 1, validation loss never fell below 0.4, while in the training from scratch approach, there is larger oscillation in the validation loss, compared to Test case 0. Additionally, the differences are more obvious in the attention maps where even in the training from scratch approach of Test case 1, the crack in the second image patch is not fully detected, contrary to the results of Test case 0. Also, similar differences are apparent in the patches after the transfer learning approach.

### 3.2 Experiments on Training - Test case 2

This Training-Test case is realized by training the models on all the imagery available, except of the images of the St. Nikolaos test site and testing of those models on this. This approach will highlight the generalization capabilities of the trained models. 39 images of those are depicting areas of masonry walls with "No cracks" while 50 images are labelled as "Cracks", totaling 89 images, were used for training while 16 and 8 images respectively were used for testing the trained models. When following the transfer learning approach InceptionResNetV2, MobileNetV3Small and MobileNetV3Large networks achieved the best accuracy scores, reaching 0.95 in 231.764, 45.558 and 47.121 seconds respectively. Learning rates were selected as follows: 0.0001. 0.000085, and 0.000085 respectively. By training from scratch the models, VGG16, ResNet50 and ResNet101 networks achieved the best accuracy scores, reaching 0.95 in 154.351, 189.391 and 337.003 seconds respectively. Learning rates were selected as follows: 0.0001. 0.000085, and 0.000085 respectively.

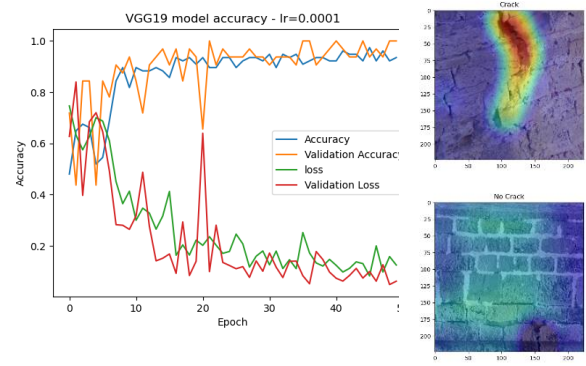

**Fig. 6 – Models' training and validation accuracy, training and validation loss and typical images from the testing datasets, demonstrating the detected crack and the predicted image class. Results for MobileNetV3Small1 are displayed on top while results for the VGG19 are displayed on the bottom.**

This approach will also highlight the generalization capabilities of the trained models. 41 images of those are depicting areas of masonry walls with "No cracks" while 36 images are labelled as "Cracks", totaling 77 images, were used for training while 14 and 22 images respectively were used for testing the trained models. When following the transfer learning approach MobileNetV3Small network achieved the best accuracy score, reaching 0.94 in 110.159 seconds. Learning rate was selected equal to 0.000085. ResNet101 follows with an accuracy score of 0.91 and total training time 362.797 seconds. By training from scratch the models, VGG19 network achieved the best accuracy score, reaching 1 in 173.208 seconds while VGG16 followed, reaching accuracy score of 0.97 in 137.379 seconds. Learning rate was selected 0.0001 for both networks.

### 3.4 Experiments on Training - Test case 4

Training-Test case 4 is the training of the models on the imagery available from the HYPERION data and testing those models on the random imagery collected by the authors and some images retrieved from the internet. This approach will also highlight the generalization capabilities of the trained models. 30 images of those are depicting areas of masonry walls with "No cracks" while 30 images are labelled as "Cracks", totaling 60 images, were used for training while 12 and 26 images respectively were used for testing the trained models. When following the transfer learning approach ResNet50 network achieved the best accuracy score, reaching 0.84 in 71.210 seconds. Learning rate was selected equal to 0.000085. DenseNet201, VGG19, MobileNetV3Small and MobileNetV3Large follow with an accuracy score of 0.81 and total training time 192.115, 68.845, 42.433 and 49.863 seconds respectively. By training from scratch the models, again, VGG19 network achieved the best accuracy score, reaching 0.97 in 173.236 seconds while VGG16 follows, reaching accuracy score of 0.95 in 138.661 seconds. Learning rate was selected 0.0001 for both networks.

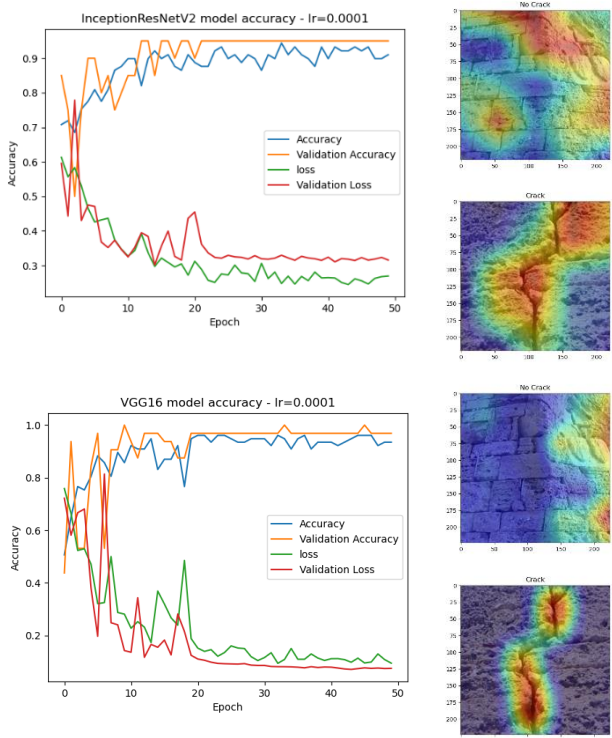

**Fig. 5 – Models' training and validation accuracy, training and validation loss and typical images from the testing datasets, demonstrating the detected crack and the predicted image class. Results for InceptionResNetV2 are displayed on top while results for the VGG16 are displayed on the bottom.**

### 3.3 Experiments on Training - Test case 3

Training-Test case 3 is the training of the models on all the imagery available, except of the images of the Naillac test site and testing of those models on this.

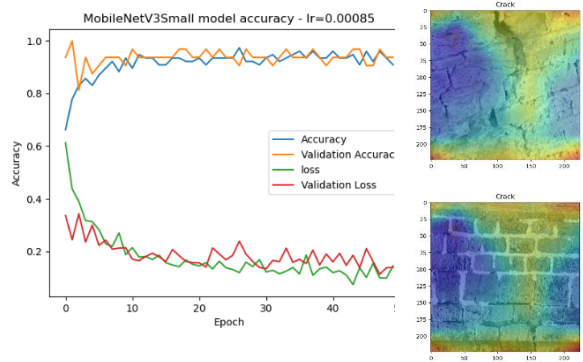

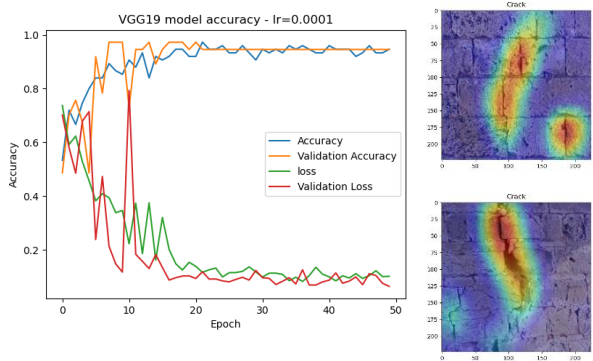

**Fig. 7 – Models' training and validation accuracy, training and validation loss and typical images from the testing datasets, demonstrating the detected crack and the predicted image class. Results for ResNet50V2 are displayed on top while results for the VGG19 are displayed on the bottom.**

### 3.5 Experiments on Training - Test case 5

This final Training-Test case 5 is realized by the training of the models on the random imagery collected by the authors and some images retrieved from the internet and testing those models on imagery available from the HYPERION data. This approach will also highlight the generalization capabilities of the trained models. 12 images of those are depicting areas of masonry walls with "No cracks" while 26 images are labelled as "Cracks", totalling 60 images, were used for training while 30 and 30 images respectively were used for testing the trained models. When following the transfer learning approach Inception-Resnet-v2 network achieved the best accuracy score, reaching only 0.66 in 227.398 seconds. Learning rate was selected equal to 0.0001. By training from scratch the models, VGG16 network achieved the best accuracy score, reaching 0.87 in 95.106 seconds while VGG19 and ResNet50 followed, with accuracy scores of 0.85 in 121.754 and 119.874 seconds respectively. Learning rate was selected 0.0001 f and 0.000085 respectively.

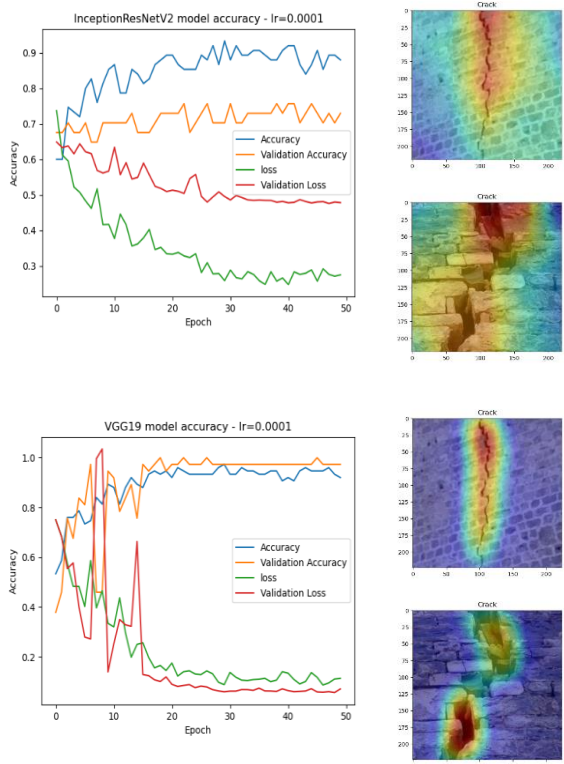

**Fig. 8 – Models' training and validation accuracy, training and validation loss and typical images from the testing datasets,**

demonstrating the detected crack and the predicted image class. Results for InceptionResNetV2 are displayed on top while results for the VGG19 are displayed on the bottom.

### 3.6 Comparative Results and Evaluation

In the figures below, results of all the performed training and test cases are presented comparatively. Fig. 9 depicts the testing accuracy of the models for each test case, after training only the additional Global Average Pooling (2D) and dense layers of the model (transfer learning approach). There, it is obvious that for the majority of the test cases, ResNet50, MobileNetV3Small and InceptionResNetV2 are achieving the best results, however, as can be seen in the Appendix Fig. A1-Fig. A6 and Tables A1-A6, localization is not correct in most of the cases. On the other hand, Fig. 10 presents testing accuracy of the same networks after being trained from scratch. There, it is clear that VGG16 and VGG19 models are outperforming the rest, while ResNet50 and ResNet101 are following. Details can be also found in the Appendix Fig. A7-Fig. A12 and Tables A7-A12. Fig. 11 and Fig. 12 comparatively present the computational time for training the described models. There, as expected, MobileNetV3Small network outperformed the rest, while MobileNetV3Large, VGG16 and VGG19 networks are following.

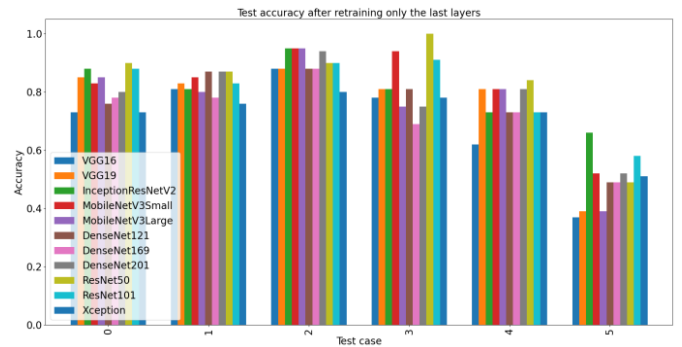

**Fig. 9 – Testing accuracy after training only the last layers**

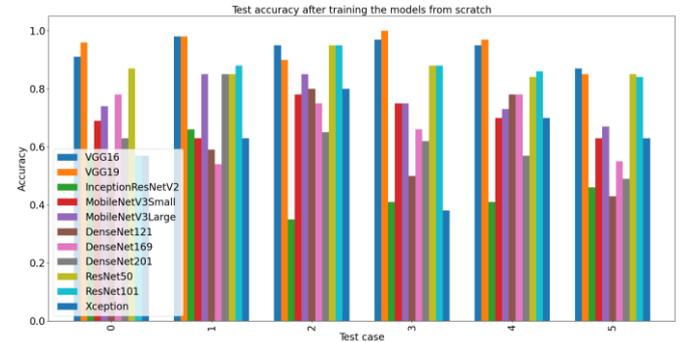

**Fig. 10 – Testing accuracy after training from scratch**

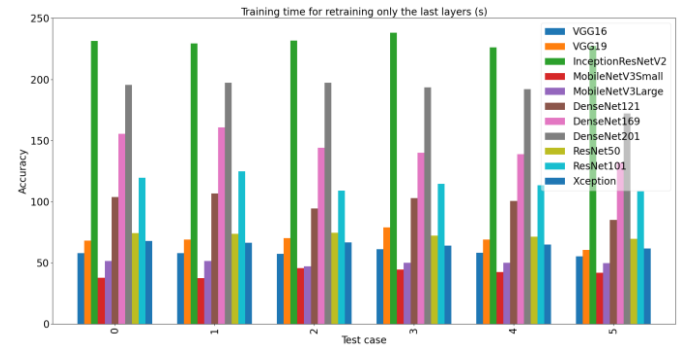

**Fig. 11 – Training time for training only the last layers**

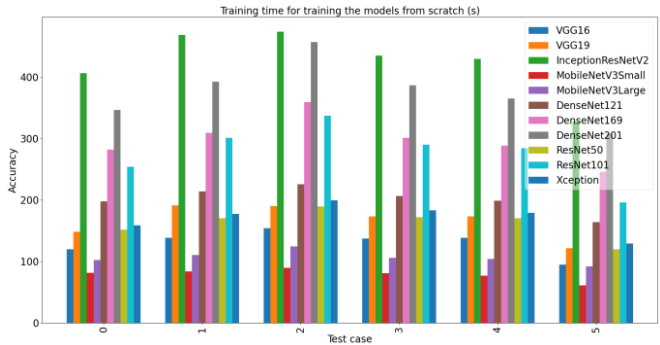

**Fig. 12 – Training time for training the model from scratch**

**3.7 Experiments on full-scale real-world image data**

To exploit the trained models over real-world data, instead of using it only with small image patches of 224x224 pixels, a sliding window approach was implemented. Sliding windows play an integral role in object detection, as they allow the localization of objects in the image. The step size of the sliding window indicates how many pixels are going to be skipped in both the (x, y) direction. Normally, loop over each and every pixel of the image is not desirable (i.e. step=1) as this would be computationally prohibitive if we were applying an image classifier at each window. Instead, the step is determined on a per-dataset basis and is tuned to give optimal performance based on your dataset of images. In the examined real-world data here, input images of 5472x3648 pixels were used and a trained model was used to predict over a sliding window of 224x224 pixels with steps of 32 pixels.

Towards that direction, for each prediction, the weights of the final dense layer of the CNN were stored in an array having the same size as the original full-resolution real-world image. To avoid issues shown in Fig. 13 (a), a fusion approach was adopted and as such, the new predictions were stacked to previous ones, enabling false positive and false negative filtering and making also the process independent of the position of the "Crack" or "Non-crack" on the image patch (Fig. 13, b).

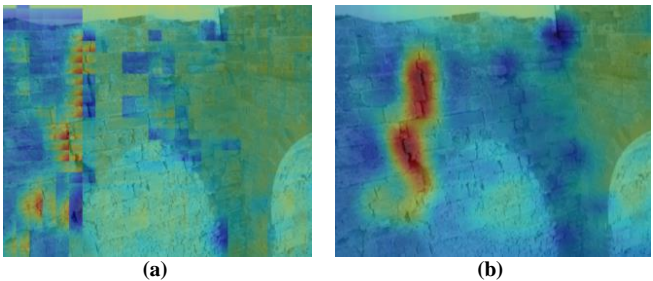

(a)                                    (b)

**Fig. 13 – The prediction of the sliding window before (a) and after (b) fusion of the results**

Figure 14 demonstrates the results of a VGG19 model trained on the CRACK-CH dataset over some real world images of cracks on stone masonry walls. Results indicate the high performance and high generalization potential of the proposed approach.

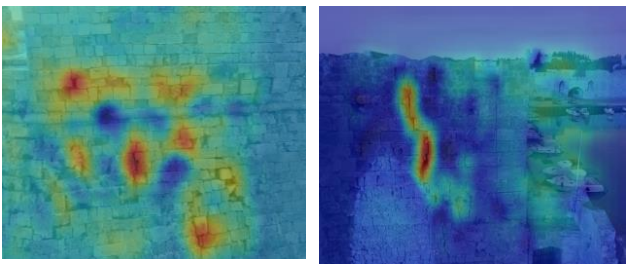

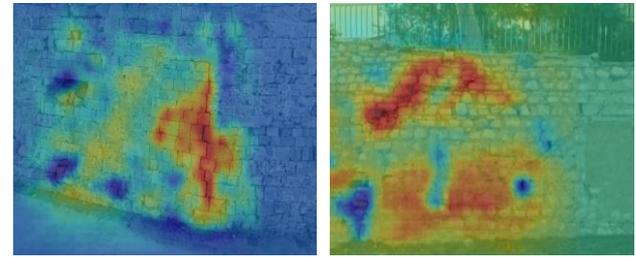

**Fig. 14 – Various results over unseen data**

The approach was also implemented in real time scenarios using cameras connected to a portable computer presenting interesting results about the real time applicability of the frameworks (Fig. 14, bottom right image).

## 4    Discussion and Concluding remarks

Taking into account the small amount of available training, validation and testing data, a typical case for Cultural Heritage applications, the CNN delivered highly accurate results in terms of image/patch classification. It is shown that even when the model is trained using data from one test site and tested over another test site, resulting accuracy is very close to 90%, similar to the accuracies achieved in the vast majority of the state of the art methods.

Considering all the tests performed, it is deduced that VGG16 and VGG19 delivered the most accurate predictions in the majority of the cases when training the models from scratch while they require some of the shortest training times. Taking into account the small number of available data for training, this success may be attributed to their small depth, 16 and 19 respectively, facilitating easier training compared to the rest of the networks which are much deeper. Inception-ResNet-v2 network provided mid accuracy levels in transfer learning tests and cracks were localized quite well, while when training the network from scratch, it was outperformed by most of the other networks and cracks weren't localized at all. Also, as observed in Fig. 11 and Fig. 12, in both approaches and for all the test cases performed it was the most expensive in terms of training time. MobileNetV3Small and Large networks provided mid accuracy levels too, in both training approaches and for all the test cases performed. However, they didn't manage to localize the cracks in the majority of the cases, even if the image patch was classified correctly as "Crack" or "No Crack". As expected, those two networks required the least time for training, compared to the rest of the networks under investigation. DenseNet121, DenseNet169 and DenseNet201 networks for all the transfer learning test cases provided less accuracy compared to the aforementioned networks while for the majority of the training from scratch test cases they resulted in higher accuracy levels compared Inception-ResNet-v2 network and lower accuracy levels compared to both MobileNetV3 architectures. They also proved to be the most time expensive networks for training, following Inception-ResNet-v2 network and Xception network, except DenseNet121. ResNet50 and ResNet101 networks achieved very high accuracy levels in the transfer learning approach while they achieved mid accuracy levels when training from scratch. Both networks managed to detect some of the cracks in the first approach while they mostly failed in the second. As expected, ResNet50 was trained faster. Finally, Xception network achieved high accuracy in the transfer learning approach and mid accuracy in the training from scratch approach. It required some of the least time to be trained in both cases while crack detection was better in the first cases. Worth noticing that after training only the last layers of the models, which is a faster process, as expected, cracks are not localized in the same detail, compared to the models trained from scratch, even if the image patch is classified correctly. This can be explained by the fact that when training the models from scratch, the extracted features are describing better the characteristics of the images used for training, thus highlighting cracks where apparent. In all the above tests it is observed that the deeper networks achieved higher accuracies and better crack localization results in the transfer learning approach. This is justified by the fact that these networks require a lot of data to be trained correctly. However, VGG16 and VGG19 proved that they can deliver better accuracy and much more accurate localization results when trained from scratch, while requiring a little time to achieve this. This can be attributed mainly to their small depth, which facilitates a correct training of the model with such a few data available.

Results suggested the high performance of the proposed approach, considering also the small numbers of epochs required for training. When training and testing was performed between different tests sites, accuracy was slightly smaller compared to the one achieved when using the shuffled data, however this was expected due to the large variations between the cracks of the two sites. Nevertheless, those results met the accuracy delivered by more complex and computationally heavy approaches, requiring a large amount of data for training.

## ACKNOWLEDGEMENTS

This work is part of the HYPERION project. HYPERION has received funding from the European Union's framework programme for Research and Innovation (HORIZON 2020) under Grant Agreement No. 821054. The content of this publication is the sole responsibility of NTUA and does not necessarily reflect the opinion of the European Union. For all figures permissions have been obtained from the owners.

APPENDIX

**PRETRAINED MODELS**

**Table A.1: Evaluation results on the 25% of the data using the pretrained models on ImageNet, after training only the additional Global Average Pooling (2D) and dense layers on the rest 75% of the data**

| model | epochs | learning rate | precision | recall | f1-score | accuracy | tr time |
|---|---|---|---|---|---|---|---|
| vgg16 | 50 | 0.0001 | 0.73 | 0.73 | 0.73 | 0.73 | 57.783 |
| vgg19 | 50 | 0.0001 | 0.87 | 0.86 | 0.85 | 0.85 | 68.080 |
| InceptionResNetV2 | 50 | 0.0001 | 0.90 | 0.88 | 0.88 | 0.88 | 231.568 |
| MobileNetV3Small | 50 | 0.0001 | 0.87 | 0.83 | 0.82 | 0.83 | 37.719 |
| MobileNetV3Large | 50 | 0.0001 | 0.87 | 0.85 | 0.85 | 0.85 | 51.524 |
| DenseNet121 | 50 | 0.0001 | 0.77 | 0.76 | 0.75 | 0.76 | 103.704 |
| DenseNet169 | 50 | 0.0001 | 0.78 | 0.78 | 0.78 | 0.78 | 155.503 |
| DenseNet201 | 50 | 0.0001 | 0.82 | 0.80 | 0.80 | 0.80 | 195.564 |
| ResNet50 | 50 | 0.000085 | 0.92 | 0.90 | 0.90 | **0.90** | 74.141 |
| ResNet101 | 50 | 0.000085 | 0.90 | 0.88 | 0.88 | 0.88 | 119.520 |
| Xception | 50 | 0.001 | 0.73 | 0.73 | 0.73 | 0.73 | 67.817 |

**Table A.2: Evaluation results on the 50% of the data using the pretrained models on ImageNet, after training only the additional Global Average Pooling (2D) and dense layers on the rest 50% of the data**

| model | epochs | learning rate | precision | recall | f1-score | accuracy | tr time |
|---|---|---|---|---|---|---|---|
| vgg16 | 50 | 0.0001 | 0.82 | 0.81 | 0.81 | 0.81 | 57.884 |
| vgg19 | 50 | 0.0001 | 0.84 | 0.83 | 0.83 | 0.83 | 68.850 |
| InceptionResNetV2 | 50 | 0.0001 | 0.86 | 0.81 | 0.81 | 0.81 | 229.487 |
| MobileNetV3Small | 50 | 0.0001 | 0.87 | 0.85 | 0.85 | 0.85 | 37.550 |
| MobileNetV3Large | 50 | 0.0001 | 0.83 | 0.80 | 0.79 | 0.80 | 51.492 |
| DenseNet121 | 50 | 0.0001 | 0.87 | 0.87 | 0.87 | **0.87** | 106.533 |
| DenseNet169 | 50 | 0.0001 | 0.79 | 0.78 | 0.77 | 0.78 | 160.684 |
| DenseNet201 | 50 | 0.0001 | 0.90 | 0.87 | 0.87 | **0.87** | 197.251 |
| ResNet50 | 50 | 0.000085 | 0.90 | 0.87 | 0.87 | **0.87** | 73.580 |
| ResNet101 | 50 | 0.000085 | 0.85 | 0.83 | 0.83 | 0.83 | 124.912 |
| Xception | 50 | 0.001 | 0.76 | 0.76 | 0.76 | 0.76 | 66.240 |

**Table A.3: Evaluation results on the St. Nikolaos test site of the data using the pretrained models on ImageNet, after training only the additional Global Average Pooling (2D) and dense layers on the rest of the data**

| model | epochs | learning rate | precision | recall | f1-score | accuracy | tr time |
|---|---|---|---|---|---|---|---|
| vgg16 | 50 | 0.0001 | 0.88 | 0.88 | 0.88 | 0.88 | 57.343 |
| vgg19 | 50 | 0.0001 | 0.88 | 0.88 | 0.88 | 0.88 | 70.236 |
| InceptionResNetV2 | 50 | 0.0001 | 0.96 | 0.95 | 0.95 | **0.95** | 231.764 |
| MobileNetV3Small | 50 | 0.00085 | 0.95 | 0.95 | 0.95 | **0.95** | 45.558 |
| MobileNetV3Large | 50 | 0.00085 | 0.96 | 0.95 | 0.95 | **0.95** | 47.121 |
| DenseNet121 | 50 | 0.0001 | 0.88 | 0.88 | 0.88 | 0.88 | 94.537 |
| DenseNet169 | 50 | 0.0001 | 0.93 | 0.88 | 0.89 | 0.88 | 144.185 |
| DenseNet201 | 50 | 0.0001 | 0.96 | 0.94 | 0.94 | 0.94 | 197.117 |
| ResNet50 | 50 | 0.000085 | 0.90 | 0.90 | 0.90 | 0.90 | 74.629 |
| ResNet101 | 50 | 0.000085 | 0.90 | 0.90 | 0.90 | 0.90 | 108.907 |
| Xception | 50 | 0.001 | 0.90 | 0.80 | 0.82 | 0.80 | 66.653 |

**Table A.4: Evaluation results on the Naillac test site of the data using the pretrained models on ImageNet, after training only the additional Global Average Pooling (2D) and dense layers on the rest of the data**

| model | epochs | learning rate | precision | recall | f1-score | accuracy | tr time |
|---|---|---|---|---|---|---|---|
| vgg16 | 50 | 0.0001 | 0.85 | 0.78 | 0.78 | 0.78 | 619.484 |
| vgg19 | 50 | 0.0001 | 0.87 | 0.81 | 0.81 | 0.81 | 500.929 |
| InceptionResNetV2 | 50 | 0.0001 | 0.87 | 0.81 | 0.81 | 0.81 | 367.272 |
| MobileNetV3Small | 50 | 0.00085 | 0.94 | 0.94 | 0.94 | 0.94 | 110.159 |
| MobileNetV3Large | 50 | 0.00085 | 0.75 | 0.75 | 0.75 | 0.75 | 140.934 |
| DenseNet121 | 50 | 0.0001 | 0.87 | 0.81 | 0.81 | 0.81 | 220.415 |
| DenseNet169 | 50 | 0.0001 | 0.82 | 0.69 | 0.67 | 0.69 | 271.657 |
| DenseNet201 | 50 | 0.0001 | 0.84 | 0.75 | 0.74 | 0.75 | 357.720 |
| ResNet50 | 50 | 0.000085 | 1 | 1 | 1 | **1** | 191.074 |
| ResNet101 | 50 | 0.000085 | 0.92 | 0.91 | 0.91 | 0.91 | 362.797 |

| Xception | 50 | 0.001 | 0.85 | 0.78 | 0.78 | 0.78 | 408.738 |
| --- | --- | --- | --- | --- | --- | --- | --- |

**Table A.5:** Evaluation results on the Internet data using the pretrained models on ImageNet, after training only the additional Global Average Pooling (2D) and dense layers on the rest of the data

| model | epochs | learning rate | precision | recall | f1-score | accuracy | tr time |
| --- | --- | --- | --- | --- | --- | --- | --- |
| vgg16 | 50 | 0.0001 | 0.74 | 0.62 | 0.64 | 0.62 | 58.285 |
| vgg19 | 50 | 0.0001 | 0.82 | 0.81 | 0.81 | 0.81 | 68.845 |
| InceptionResNetV2 | 50 | 0.0001 | 0.71 | 0.73 | 0.69 | 0.73 | 226.108 |
| MobileNetV3Small | 50 | 0.00085 | 0.80 | 0.81 | 0.80 | 0.81 | 42.433 |
| MobileNetV3Large | 50 | 0.00085 | 0.82 | 0.81 | 0.81 | 0.81 | 49.863 |
| DenseNet121 | 50 | 0.0001 | 0.71 | 0.73 | 0.71 | 0.73 | 100.599 |
| DenseNet169 | 50 | 0.0001 | 0.73 | 0.73 | 0.73 | 0.73 | 138.728 |
| DenseNet201 | 50 | 0.0001 | 0.81 | 0.81 | 0.79 | 0.81 | 192.115 |
| ResNet50 | 50 | 0.000085 | 0.87 | 0.84 | 0.84 | **0.84** | 71.210 |
| ResNet101 | 50 | 0.000085 | 0.71 | 0.73 | 0.71 | 0.73 | 113.310 |
| Xception | 50 | 0.001 | 0.72 | 0.73 | 0.72 | 0.73 | 64.893 |

**Table A.6:** Evaluation results on the HYPERION data using the pretrained models on ImageNet, after training only the additional Global Average Pooling (2D) and dense layers on the internet data

| model | epochs | learning rate | precision | recall | f1-score | accuracy | tr time |
| --- | --- | --- | --- | --- | --- | --- | --- |
| vgg16 | 50 | 0.0001 | 0.14 | 0.37 | 0.20 | 0.37 | 55.387 |
| vgg19 | 50 | 0.0001 | 0.77 | 0.39 | 0.23 | 0.39 | 60.563 |
| InceptionResNetV2 | 50 | 0.0001 | 0.72 | 0.66 | 0.66 | **0.66** | 227.398 |
| MobileNetV3Small | 50 | 0.00085 | 0.74 | 0.52 | 0.48 | 0.52 | 41.848 |
| MobileNetV3Large | 50 | 0.00085 | 0.48 | 0.39 | 0.31 | 0.39 | 49.676 |
| DenseNet121 | 50 | 0.0001 | 0.68 | 0.49 | 0.45 | 0.49 | 84.917 |
| DenseNet169 | 50 | 0.0001 | 0.72 | 0.49 | 0.44 | 0.49 | 131.652 |
| DenseNet201 | 50 | 0.0001 | 0.59 | 0.52 | 0.52 | 0.52 | 172.098 |
| ResNet50 | 50 | 0.000085 | 0.65 | 0.49 | 0.46 | 0.49 | 69.546 |
| ResNet101 | 50 | 0.000085 | 0.71 | 0.58 | 0.57 | 0.58 | 108.361 |
| Xception | 50 | 0.001 | 0.64 | 0.51 | 0.49 | 0.51 | 61.722 |

**Fig. A.1:** Evaluation results on the 50% of the data using the pretrained models on ImageNet, after training only the additional Global Average Pooling (2D) and dense layers on the rest 50% of the data

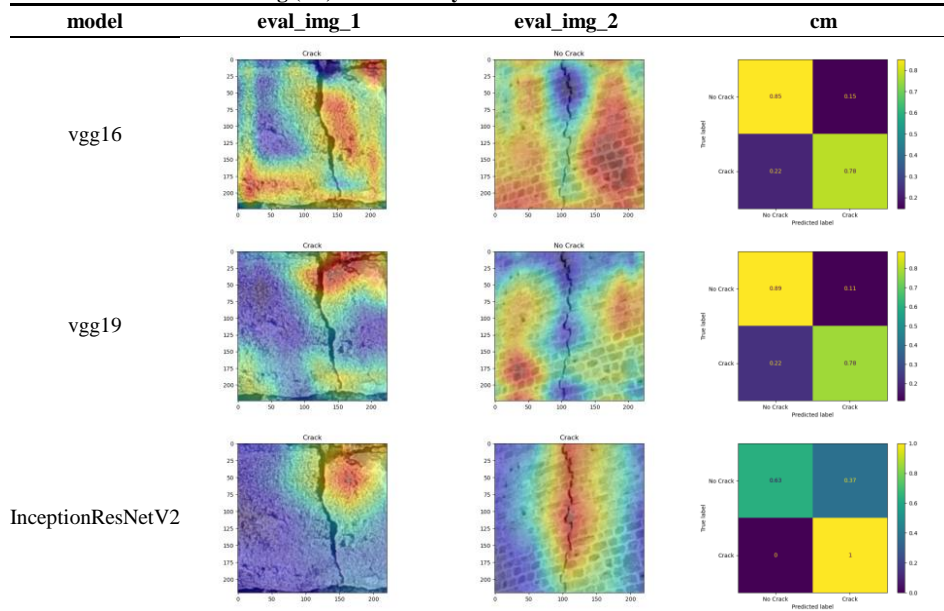

| model | eval_img_1 | eval_img_2 | cm |
| --- | --- | --- | --- |
| vgg16 | | | |
| vgg19 | | | |
| InceptionResNetV2 | | | |

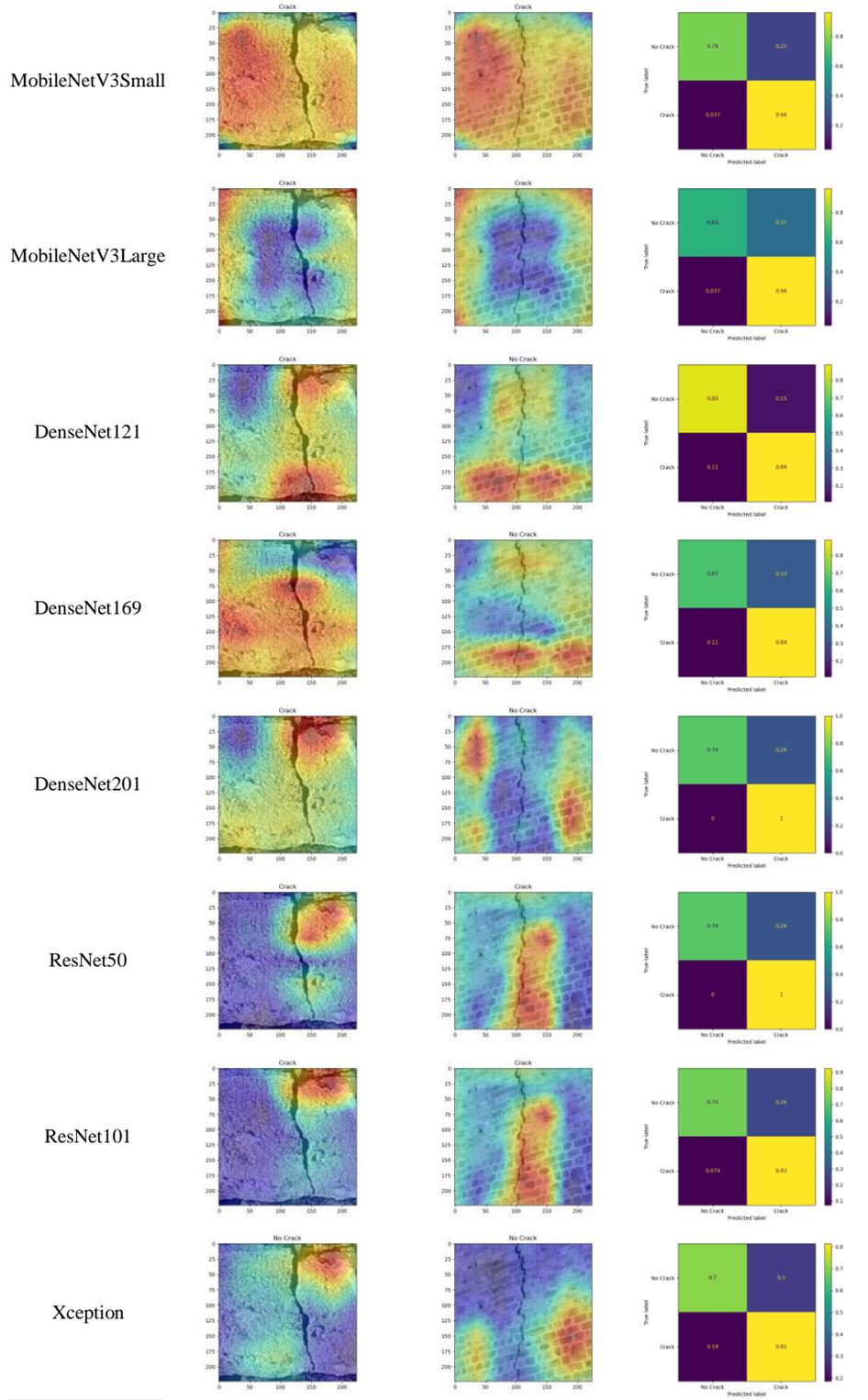

**Fig. A.2: Evaluation results on the 25% of the data using the pretrained models on ImageNet, after training only the additional Global Average Pooling (2D) and dense layers on the rest 75% of the data**

| model | eval_img_1 | eval_img_2 | cm |

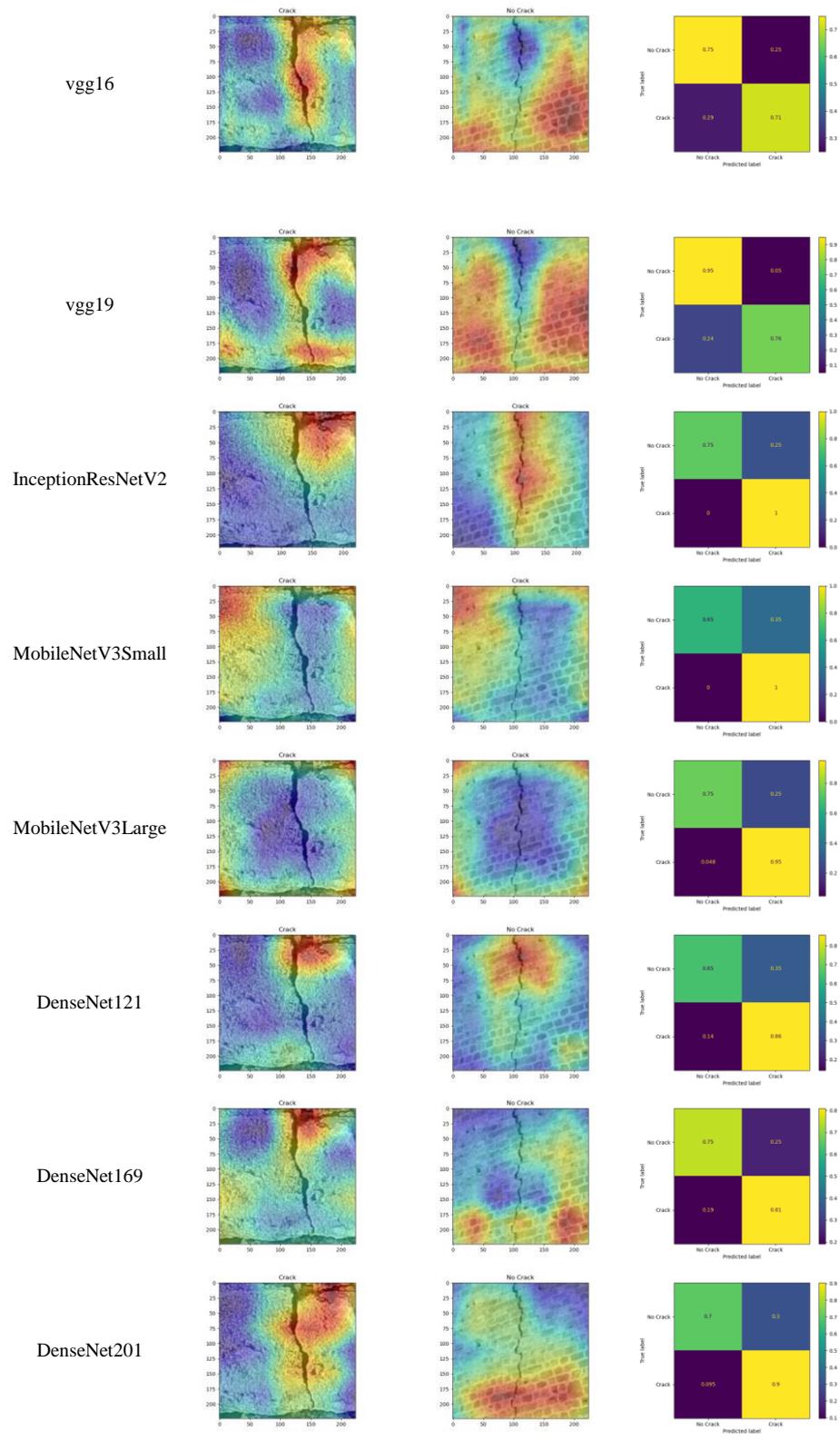

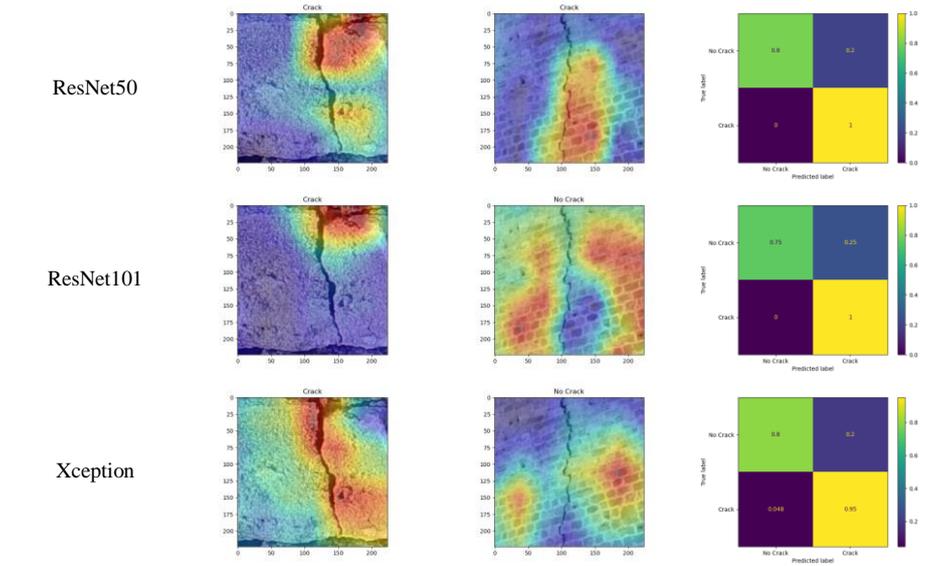

**Fig. A.3: Evaluation results on the Naillac test site of the data using the pretrained models on ImageNet, after training only the additional Global Average Pooling (2D) and dense layers on the rest of the data**

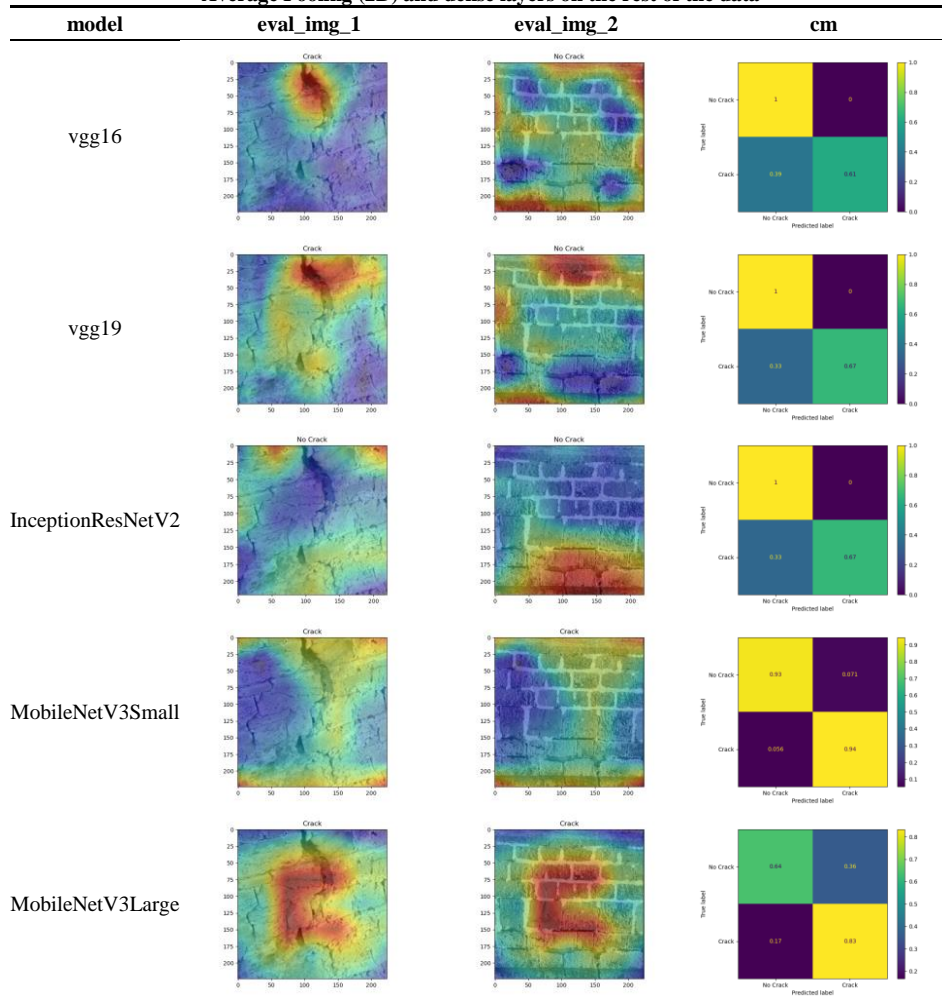

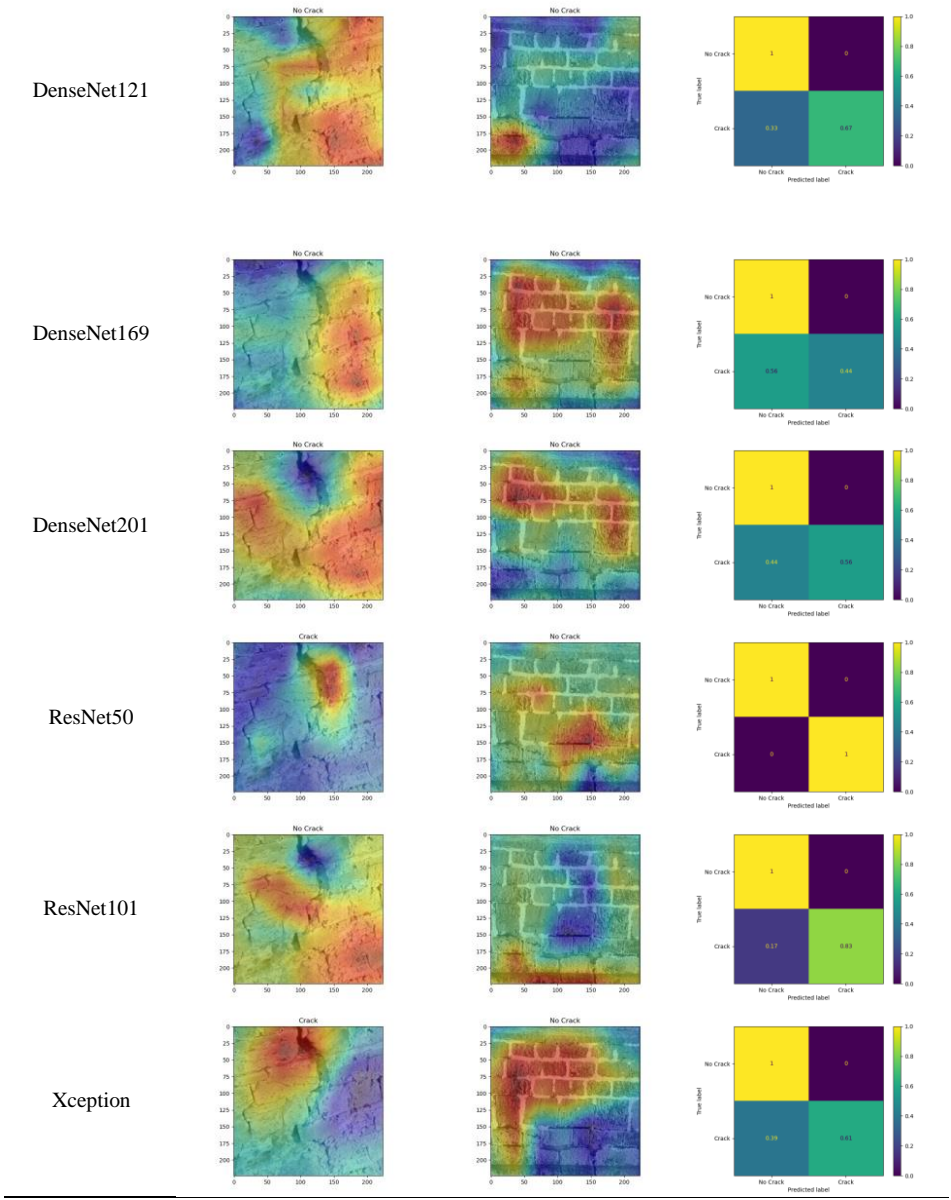

**Fig. A.4:** Evaluation results on the St. Nikolaos test site of the data using the pretrained models on ImageNet, after training only the additional Global Average Pooling (2D) and dense layers on the rest of the data

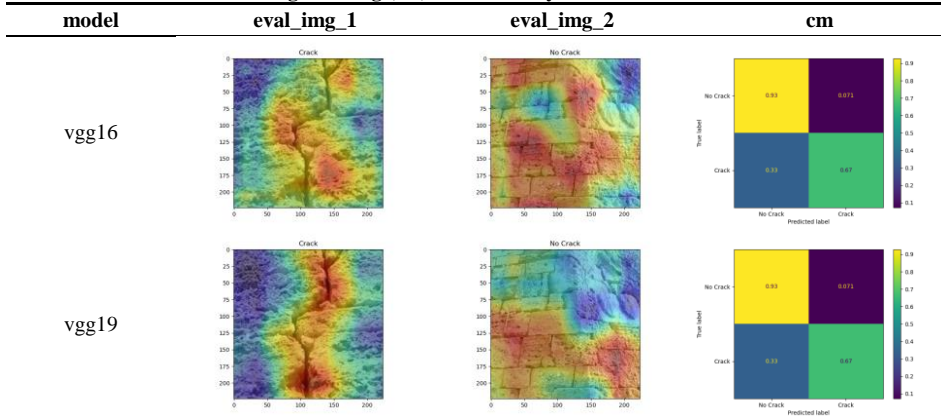

InceptionResNetV2
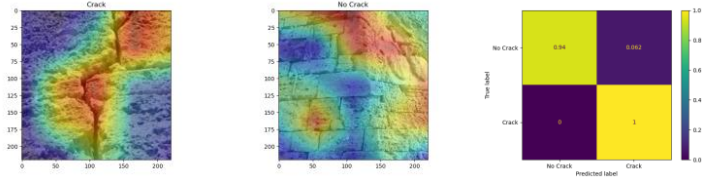

MobileNetV3Small
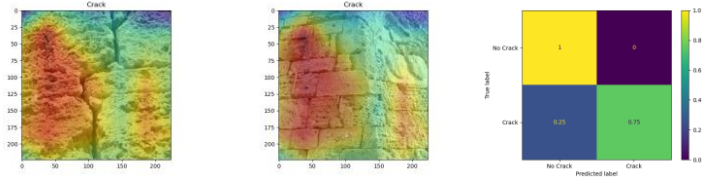

MobileNetV3Large
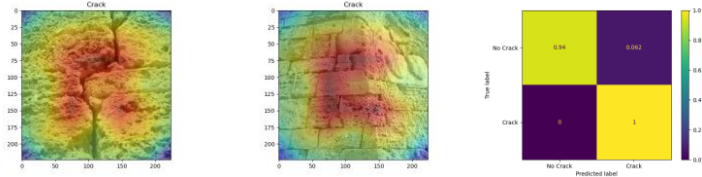

DenseNet121
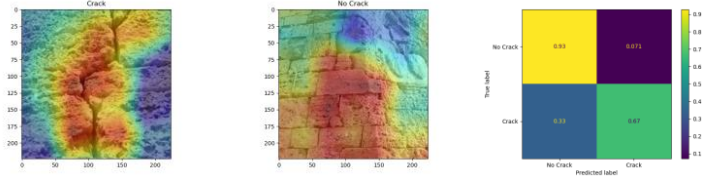

DenseNet169
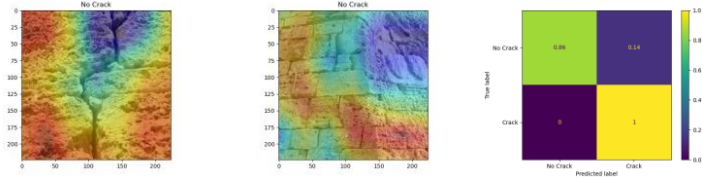

DenseNet201
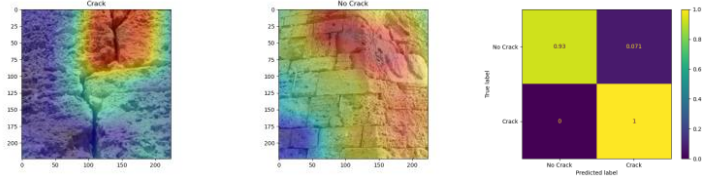

ResNet50
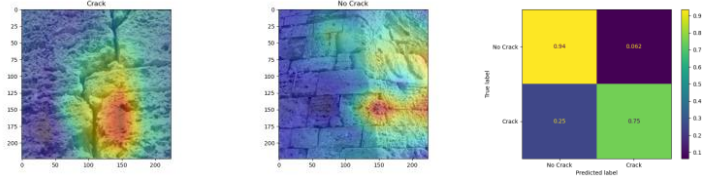

ResNet101
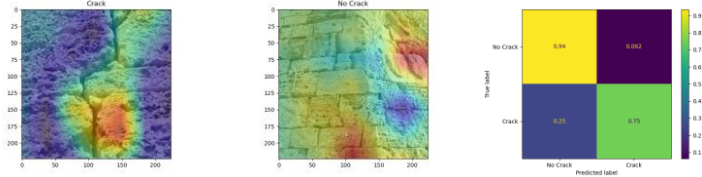

| Xception | 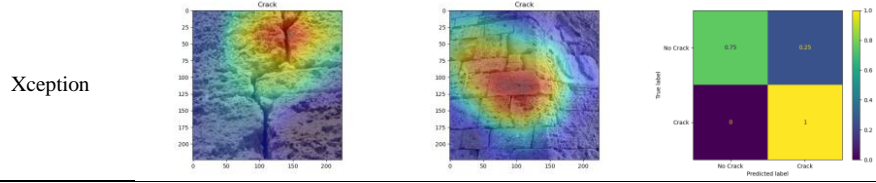 |
|----------|----------------------|

**Fig. A.5:** Evaluation results on the Internet test site of the data using the pretrained models on ImageNet, after training only the additional Global Average Pooling (2D) and dense layers on the rest of the data

| model | eval_img_1 | eval_img_2 | cm |
|-------|------------|------------|-----|
| vgg16 | | | |
| vgg19 | | | |
| InceptionResNetV2 | | | |
| MobileNetV3Small | | | |
| MobileNetV3Large | | | |
| DenseNet121 | | | |
| DenseNet169 | | | |

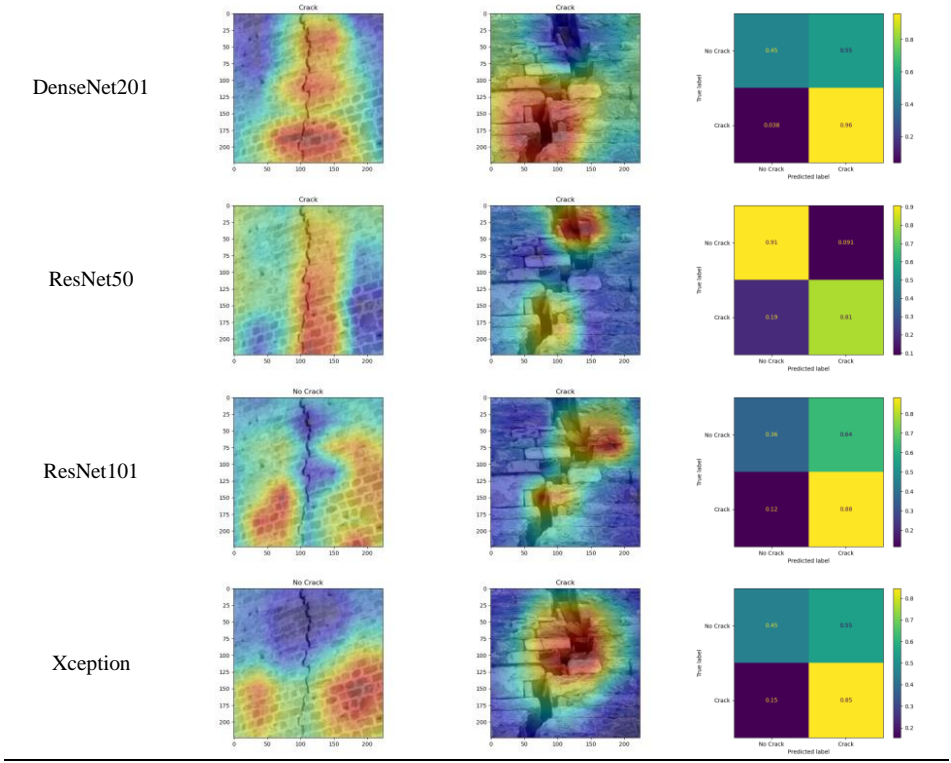

**Fig. A.6:** Evaluation results on the Rest test sites of the data using the pretrained models on ImageNet, after training only the additional Global Average Pooling (2D) and dense layers on the internet data

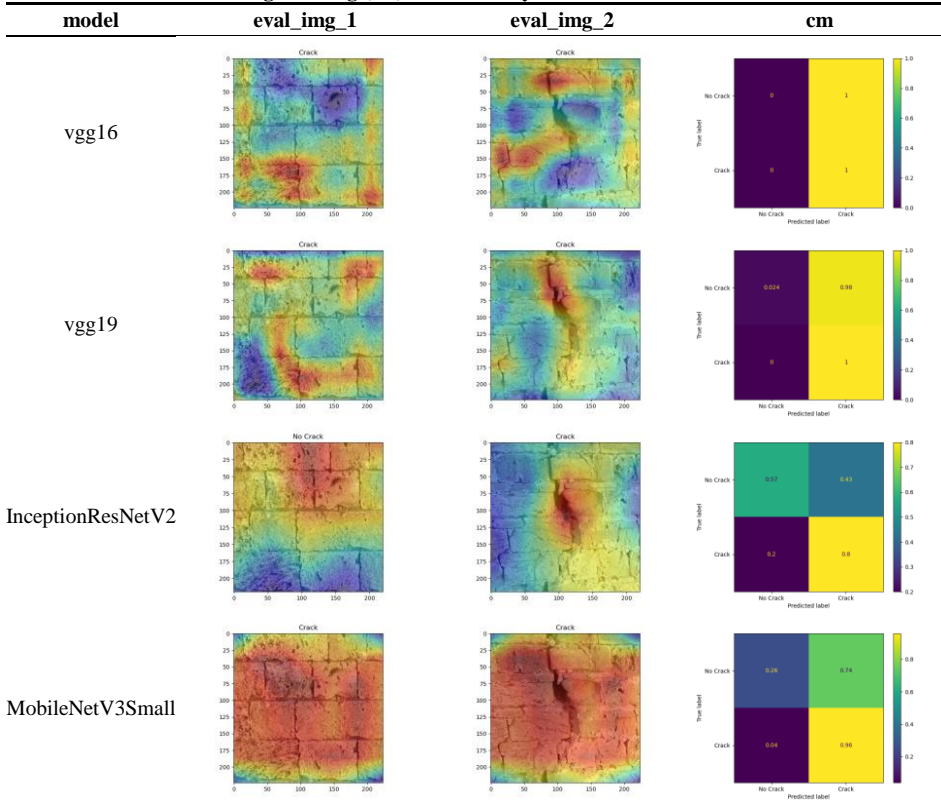

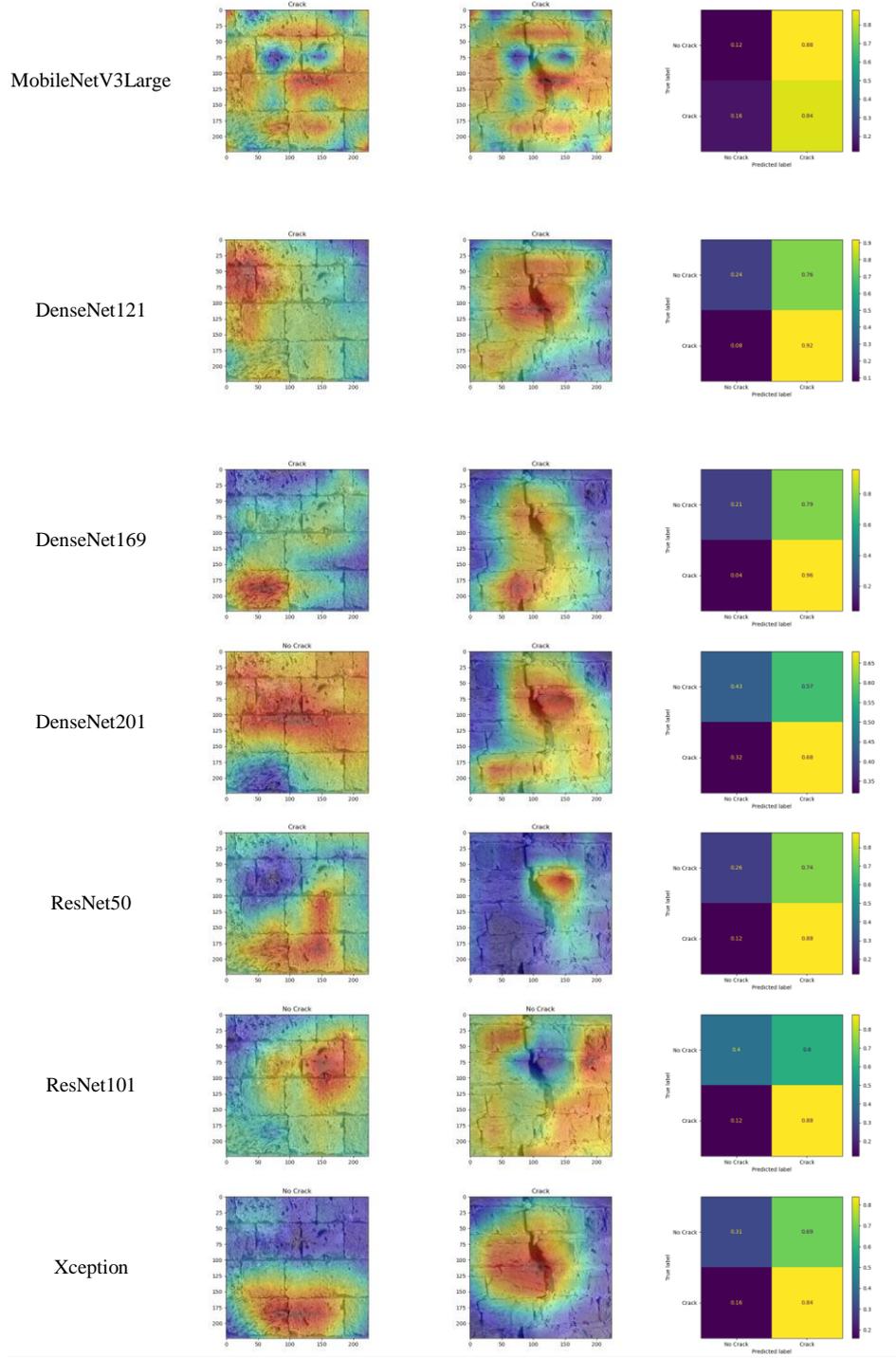

**TRAINING FROM SCRATCH**

**Table A.7: Evaluation results on the HYPERION data using the pretrained models on ImageNet, after training only the additional Global Average Pooling (2D) and dense layers on the internet data**

| model | epochs | learning rate | precision | recall | f1-score | accuracy | tr time |
|---|---|---|---|---|---|---|---|
| vgg16 | 50 | 0.0001 | 0.90 | 0.87 | 0.87 | **0.87** | 95.106 |
| vgg19 | 50 | 0.0001 | 0.89 | 0.85 | 0.85 | 0.85 | 121.754 |
| InceptionResNetV2 | 50 | 0.0001 | 0.78 | 0.46 | 0.37 | 0.46 | 327.446 |

| model | epochs | learning rate | precision | recall | f1-score | accuracy | tr time |
|---|---|---|---|---|---|---|---|
| MobileNetV3Small | 50 | 0.00085 | 0.39 | 0.63 | 0.48 | 0.63 | 61.133 |
| MobileNetV3Large | 50 | 0.00085 | 0.78 | 0.67 | 0.58 | 0.67 | 92.152 |
| DenseNet121 | 50 | 0.0001 | 0.50 | 0.43 | 0.43 | 0.43 | 164.187 |
| DenseNet169 | 50 | 0.0001 | 0.53 | 0.55 | 0.54 | 0.55 | 246.038 |
| DenseNet201 | 50 | 0.0001 | 0.45 | 0.49 | 0.47 | 0.49 | 307.981 |
| ResNet50 | 50 | 0.000085 | 0.88 | 0.785 | 0.84 | 0.85 | 119.874 |
| ResNet101 | 50 | 0.000085 | 0.84 | 0.84 | 0.83 | 0.84 | 195.752 |
| Xception | 50 | 0.001 | 0.39 | 0.63 | 0.48 | 0.63 | 129.368 |

**Table A.8:** Evaluation results on the Internet data using the pretrained models on ImageNet, after training only the additional Global Average Pooling (2D) and dense layers on the HYPERION data

| model | epochs | learning rate | precision | recall | f1-score | accuracy | tr time |
|---|---|---|---|---|---|---|---|
| vgg16 | 50 | 0.0001 | 0.95 | 0.95 | 0.95 | 0.95 | 138.661 |
| vgg19 | 50 | 0.0001 | 0.98 | 0.97 | 0.97 | **0.97** | 173.236 |
| InceptionResNetV2 | 50 | 0.0001 | 0.62 | 0.41 | 0.38 | 0.41 | 429.764 |
| MobileNetV3Small | 50 | 0.00085 | 0.85 | 0.70 | 0.71 | 0.70 | 76.970 |
| MobileNetV3Large | 50 | 0.00085 | 0.82 | 0.73 | 0.74 | 0.73 | 104.207 |
| DenseNet121 | 50 | 0.0001 | 0.79 | 0.78 | 0.75 | 0.78 | 198.619 |
| DenseNet169 | 50 | 0.0001 | 0.82 | 0.78 | 0.79 | 0.78 | 288.190 |
| DenseNet201 | 50 | 0.0001 | 0.82 | 0.57 | 0.56 | 0.57 | 365.423 |
| ResNet50 | 50 | 0.000085 | 0.90 | 0.84 | 0.84 | 0.84 | 170.111 |
| ResNet101 | 50 | 0.000085 | 0.88 | 0.86 | 0.87 | 0.86 | 284.302 |
| Xception | 50 | 0.001 | 0.49 | 0.70 | 0.58 | 0.70 | 178.950 |

**Table A.9:** Evaluation results on the Naillac test site of the data using the pretrained models on ImageNet, after training only the additional Global Average Pooling (2D) and dense layers on the rest of the data

| model | epochs | learning rate | precision | recall | f1-score | accuracy | tr time |
|---|---|---|---|---|---|---|---|
| vgg16 | 50 | 0.0001 | 0.97 | 0.97 | 0.97 | 0.97 | 137.379 |
| vgg19 | 50 | 0.0001 | 1.00 | 1.00 | 1.00 | **1.00** | 173.208 |
| InceptionResNetV2 | 50 | 0.0001 | 0.42 | 0.41 | 0.36 | 0.41 | 435.101 |
| MobileNetV3Small | 50 | 0.00085 | 0.75 | 0.75 | 0.75 | 0.75 | 80.744 |
| MobileNetV3Large | 50 | 0.00085 | 0.84 | 0.75 | 0.74 | 0.75 | 106.098 |
| DenseNet121 | 50 | 0.0001 | 0.45 | 0.50 | 0.45 | 0.50 | 206.164 |
| DenseNet169 | 50 | 0.0001 | 0.71 | 0.66 | 0.65 | 0.66 | 301.307 |
| DenseNet201 | 50 | 0.0001 | 0.73 | 0.62 | 0.60 | 0.62 | 386.850 |
| ResNet50 | 50 | 0.000085 | 0.90 | 0.88 | 0.87 | 0.88 | 172.175 |
| ResNet101 | 50 | 0.000085 | 0.90 | 0.88 | 0.88 | 0.88 | 290.201 |
| Xception | 50 | 0.001 | 0.18 | 0.38 | 0.24 | 0.38 | 183.201 |

**Table A.10:** Evaluation results on the St. Nikolaos test site of the data using the pretrained models on ImageNet, after training only the additional Global Average Pooling (2D) and dense layers on the rest of the data

| model | epochs | learning rate | precision | recall | f1-score | accuracy | tr time |
|---|---|---|---|---|---|---|---|
| vgg16 | 50 | 0.0001 | 0.95 | 0.95 | 0.95 | **0.95** | 154.351 |
| vgg19 | 50 | 0.0001 | 0.91 | 0.90 | 0.89 | 0.90 | 190.256 |
| InceptionResNetV2 | 50 | 0.0001 | 0.68 | 0.35 | 0.37 | 0.35 | 473.821 |
| MobileNetV3Small | 50 | 0.0001 | 0.85 | 0.78 | 0.78 | 0.78 | 89.672 |
| MobileNetV3Large | 50 | 0.001 | 0.87 | 0.85 | 0.81 | 0.85 | 124.532 |
| DenseNet121 | 50 | 0.0001 | 0.64 | 0.80 | 0.71 | 0.80 | 225.506 |
| DenseNet169 | 50 | 0.001 | 0.77 | 0.75 | 0.76 | 0.75 | 359.552 |
| DenseNet201 | 50 | 0.0001 | 0.61 | 0.65 | 0.63 | 0.65 | 457.226 |
| ResNet50 | 50 | 0.000085 | 0.95 | 0.95 | 0.95 | **0.95** | 189.391 |
| ResNet101 | 50 | 0.000085 | 0.96 | 0.95 | 0.95 | **0.95** | 337.003 |
| Xception | 50 | 0.001 | 0.64 | 0.80 | 0.71 | 0.80 | 199.586 |

**Table A.11:** Evaluation results on the 25% of the data using the pretrained models on ImageNet, after training only the additional Global Average Pooling (2D) and dense layers on the rest 75% of the data

| model | epochs | learning rate | precision | recall | f1-score | accuracy | tr time |
|---|---|---|---|---|---|---|---|
| vgg16 | 50 | 0.0001 | 0.98 | 0.98 | 0.98 | **0.98** | 138.383 |
| vgg19 | 50 | 0.0001 | 0.98 | 0.98 | 0.98 | **0.98** | 191.179 |
| InceptionResNetV2 | 50 | 0.0001 | 0.68 | 0.66 | 0.64 | 0.66 | 468.812 |
| MobileNetV3Small | 50 | 0.0001 | 0.65 | 0.63 | 0.62 | 0.63 | 83.795 |
| MobileNetV3Large | 50 | 0.0001 | 0.87 | 0.85 | 0.85 | 0.85 | 110.798 |
| DenseNet121 | 50 | 0.0001 | 0.63 | 0.59 | 0.56 | 0.59 | 214.289 |
| DenseNet169 | 50 | 0.0001 | 0.54 | 0.54 | 0.51 | 0.54 | 309.275 |
| DenseNet201 | 50 | 0.00001 | 0.86 | 0.85 | 0.85 | 0.85 | 392.268 |
| ResNet50 | 50 | 0.000085 | 0.87 | 0.85 | 0.85 | 0.85 | 170.594 |
| ResNet101 | 50 | 0.000085 | 0.89 | 0.88 | 0.88 | 0.88 | 301.312 |
| Xception | 50 | 0.000085 | 0.79 | 0.63 | 0.58 | 0.63 | 177.304 |

**Table A.12:** Evaluation results on the 50% of the data using the pretrained models on ImageNet, after training only the additional Global Average Pooling (2D) and dense layers on the rest 50% of the data

| model | epochs | learning rate | precision | recall | f1-score | accuracy | tr time |
|---|---|---|---|---|---|---|---|
| vgg16 | 50 | 0.0001 | 0.92 | 0.91 | 0.91 | 0.91 | 119.763 |
| vgg19 | 50 | 0.0001 | 0.97 | 0.96 | 0.96 | **0.96** | 148.327 |
| InceptionResNetV2 | 50 | 0.0001 | 0.60 | 0.57 | 0.55 | 0.57 | 406.521 |
| MobileNetV3Small | 50 | 0.0001 | 0.69 | 0.69 | 0.68 | 0.69 | 81.259 |
| MobileNetV3Large | 50 | 0.0001 | 0.80 | 0.74 | 0.73 | 0.74 | 102.366 |
| DenseNet121 | 50 | 0.0001 | 0.60 | 0.57 | 0.55 | 0.57 | 197.759 |
| DenseNet169 | 50 | 0.00001 | 0.78 | 0.78 | 0.78 | 0.78 | 281.790 |
| DenseNet201 | 50 | 0.00001 | 0.66 | 0.63 | 0.61 | 0.63 | 346.763 |
| ResNet50 | 50 | 0.000085 | 0.88 | 0.87 | 0.87 | 0.87 | 151.515 |
| ResNet101 | 50 | 0.000085 | 0.57 | 0.57 | 0.57 | 0.57 | 253.925 |
| Xception | 50 | 0.000085 | 0.58 | 0.57 | 0.56 | 0.57 | 158.754 |

**Fig. A.7:** Evaluation results on the 50% of the data using the pretrained models on ImageNet, after training only the additional Global Average Pooling (2D) and dense layers on the rest 50% of the data

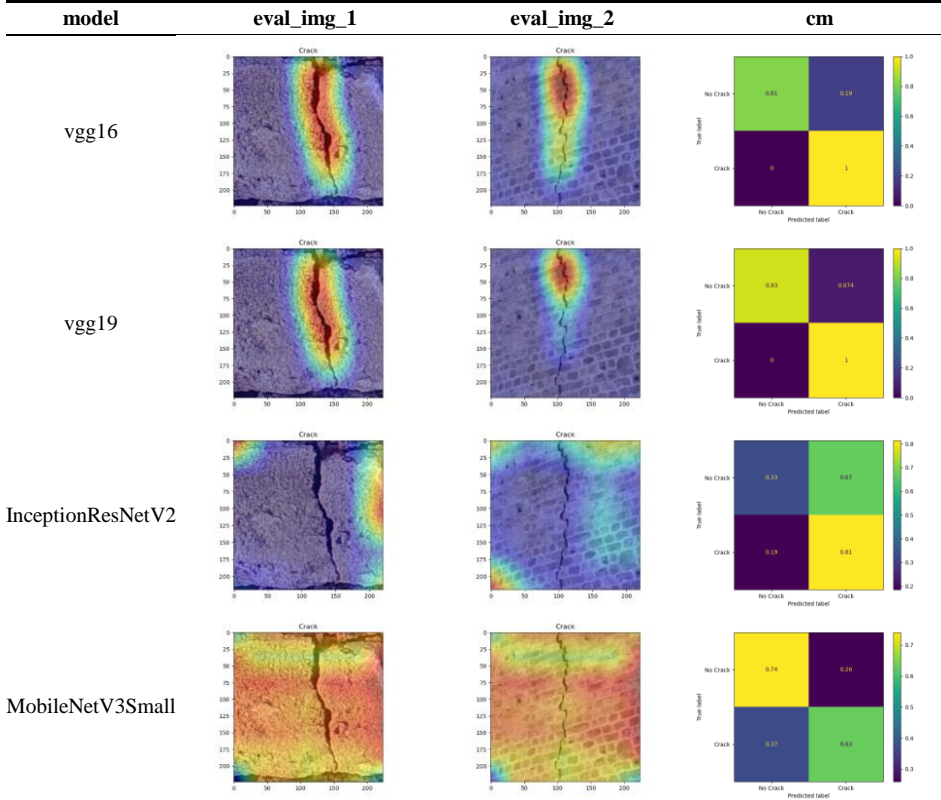

| model | eval_img_1 | eval_img_2 | cm |
|---|---|---|---|
| vgg16 | | | |
| vgg19 | | | |
| InceptionResNetV2 | | | |
| MobileNetV3Small | | | |

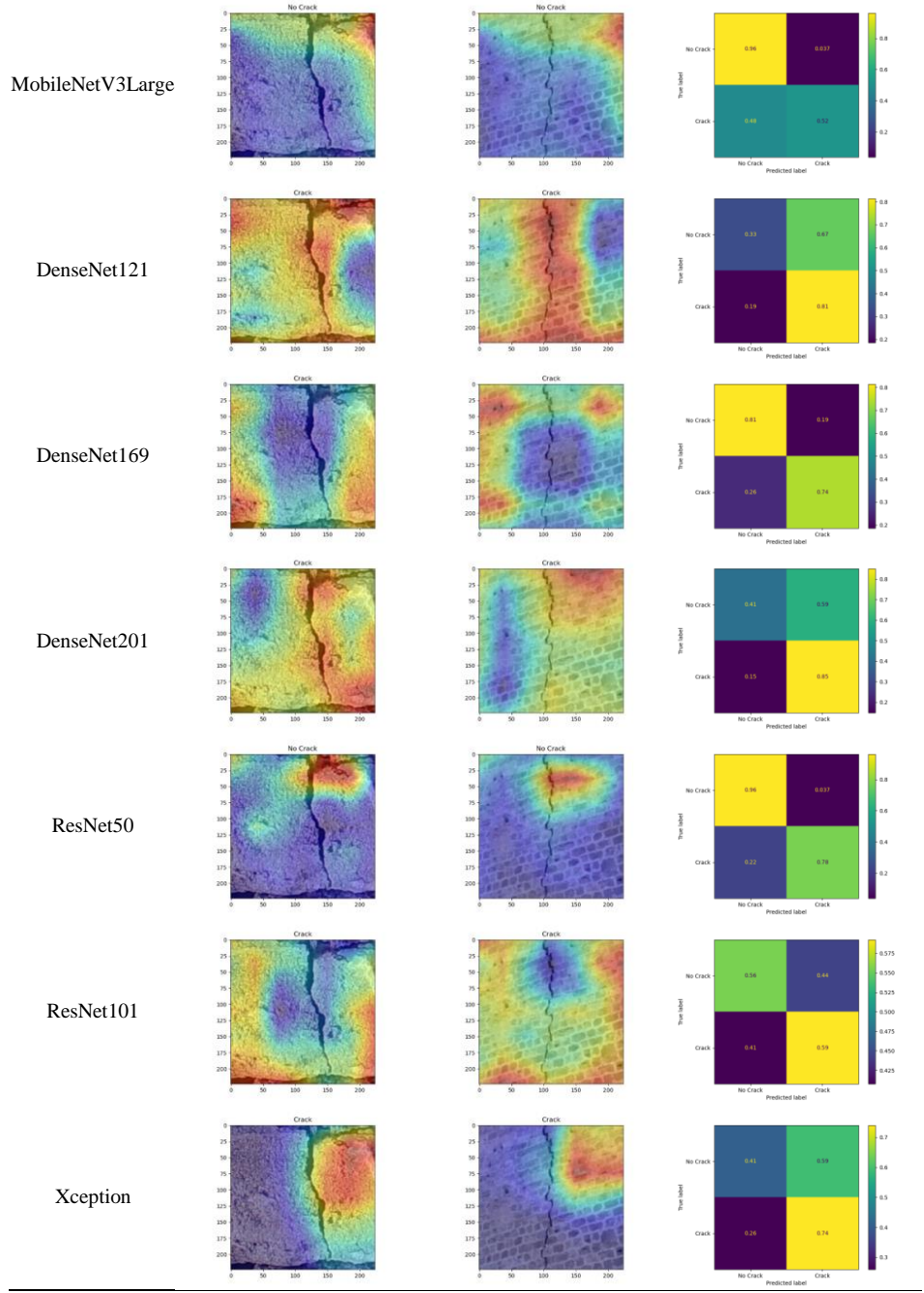

**Fig. A.8: Evaluation results on the 25% of the data using the pretrained models on ImageNet, after training only the additional Global Average Pooling (2D) and dense layers on the rest 75% of the data**

| model | eval_img_1 | eval_img_2 | cm |
|---|---|---|---|
| vgg16 | | | |

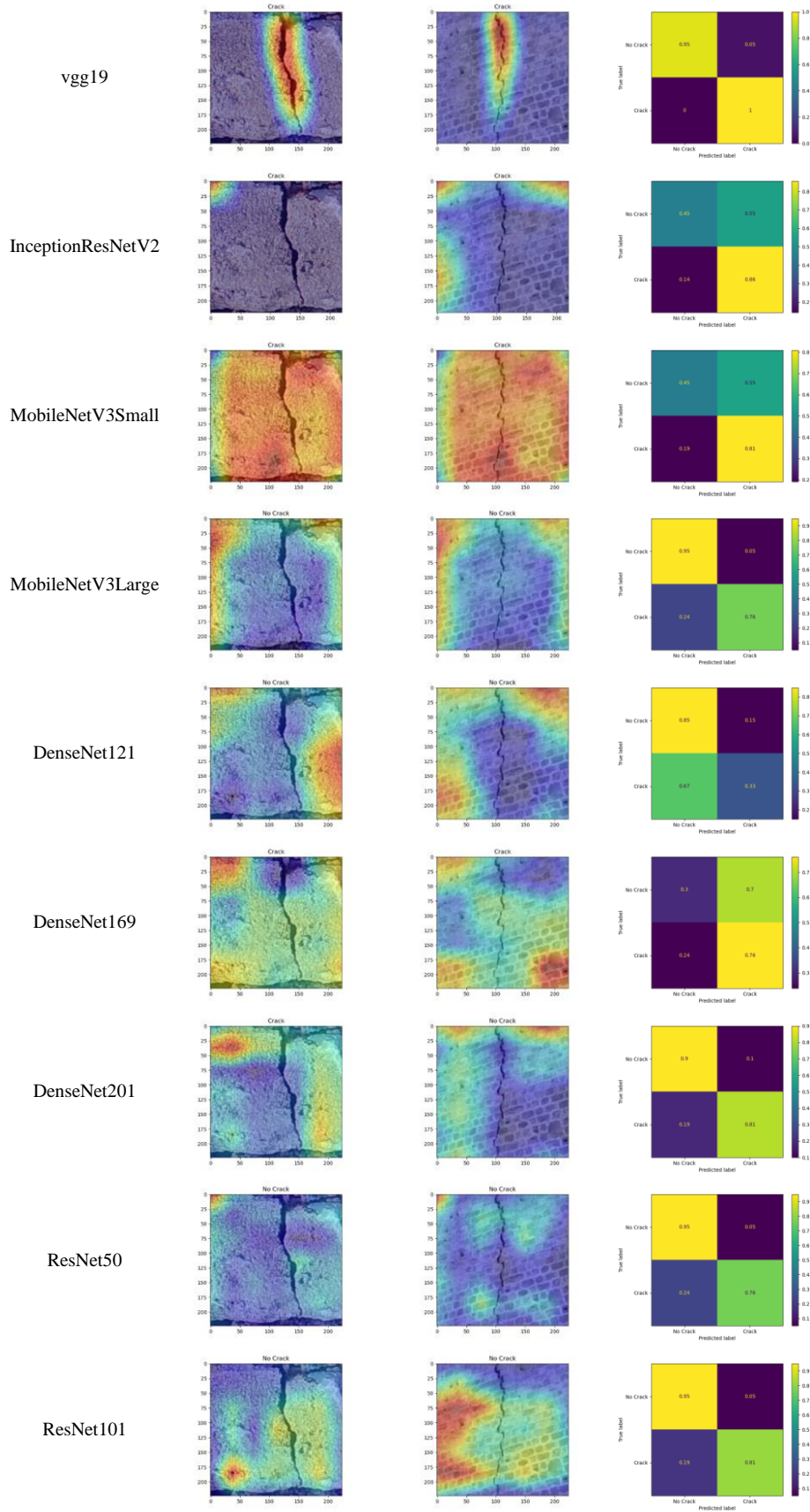

| | | |
|---|---|---|
| Xception | 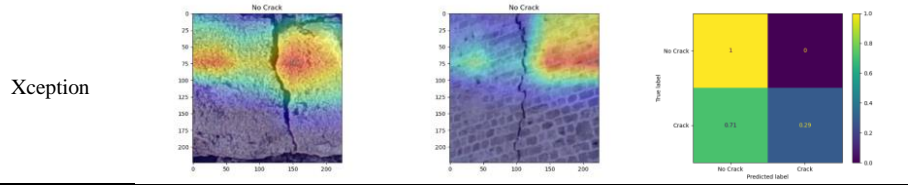 | |

**Fig. A.9:** Evaluation results on the Naillac test site of the data using the pretrained models on ImageNet, after training only the additional Global Average Pooling (2D) and dense layers on the rest of the data

| model | eval_img_1 | eval_img_2 | cm |
|---|---|---|---|
| vgg16 | | | |
| vgg19 | | | |
| InceptionResNetV2 | | | |
| MobileNetV3Small | | | |
| MobileNetV3Large | | | |
| DenseNet121 | | | |
| DenseNet169 | | | |

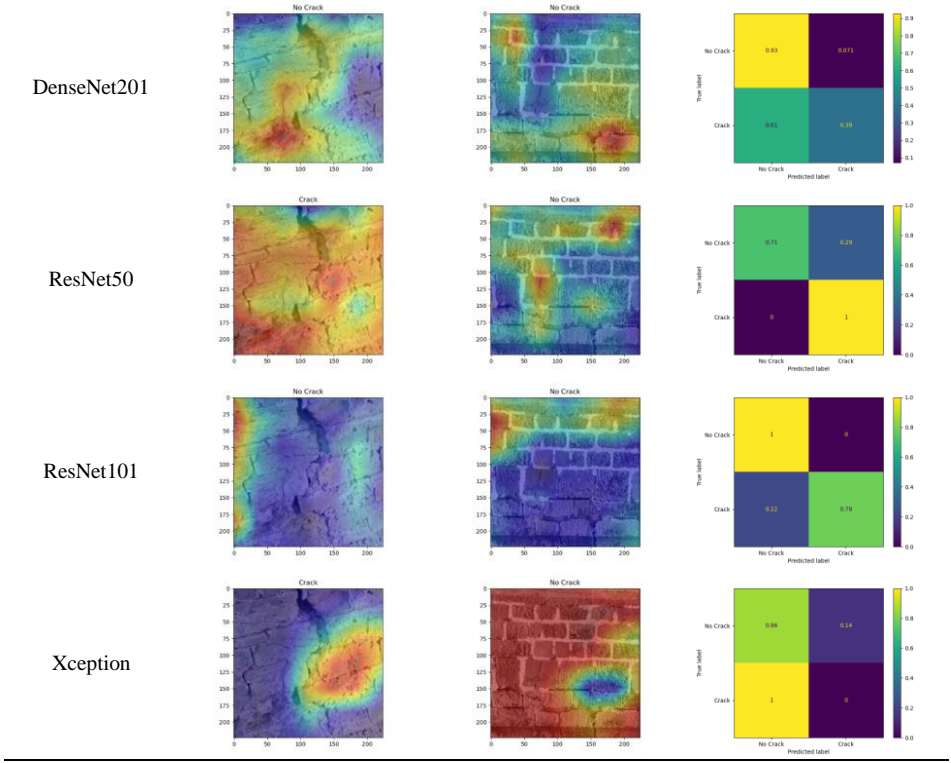

Fig. A.10: Evaluation results on the St. Nikolaos test site of the data using the pretrained models on ImageNet, after training only the additional Global Average Pooling (2D) and dense layers on the rest of the data

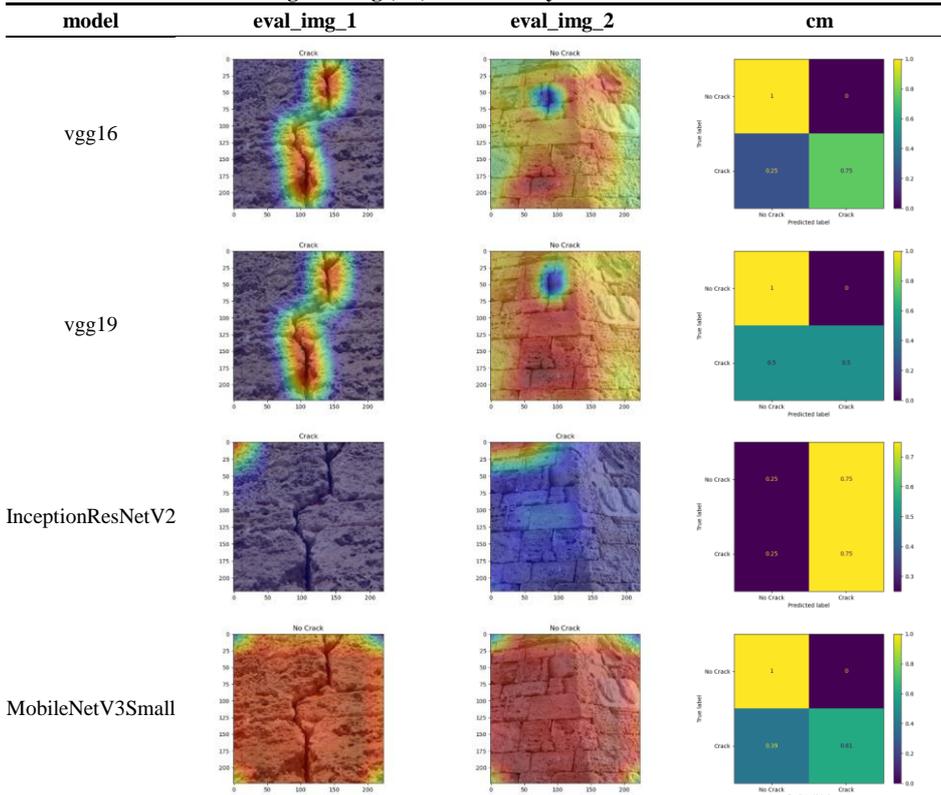

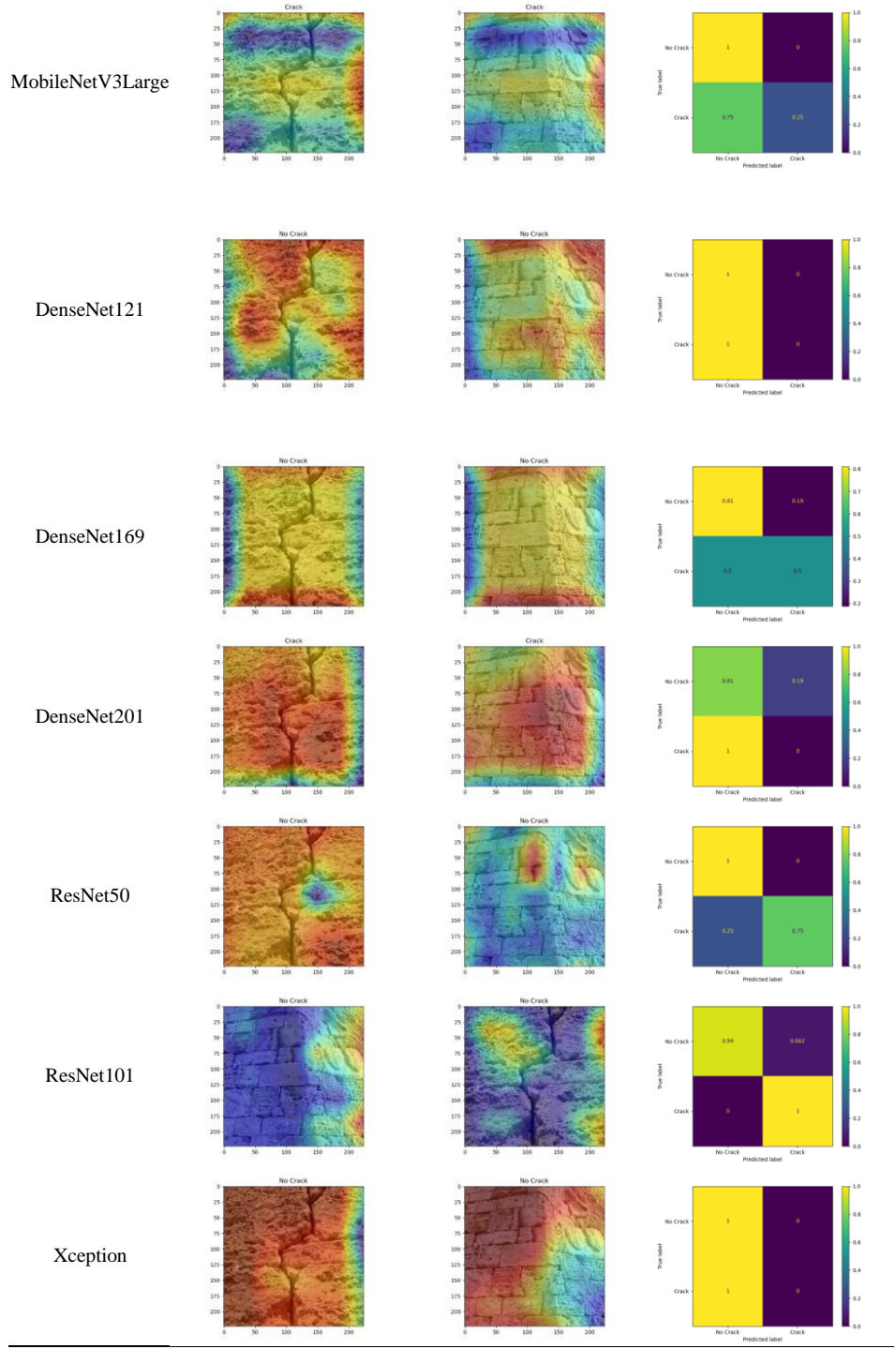

**Fig. A.11:** Evaluation results on the Internet test site of the data using the pretrained models on ImageNet, after training only the additional Global Average Pooling (2D) and dense layers on the rest of the data

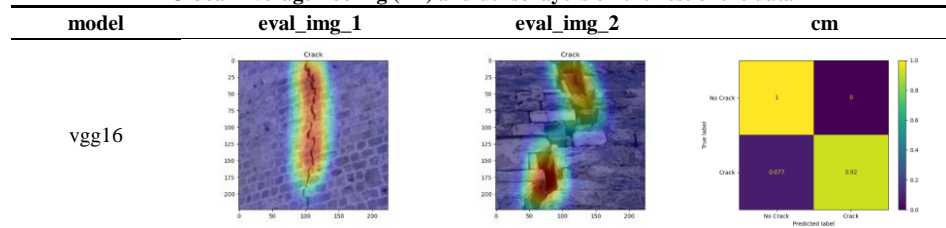

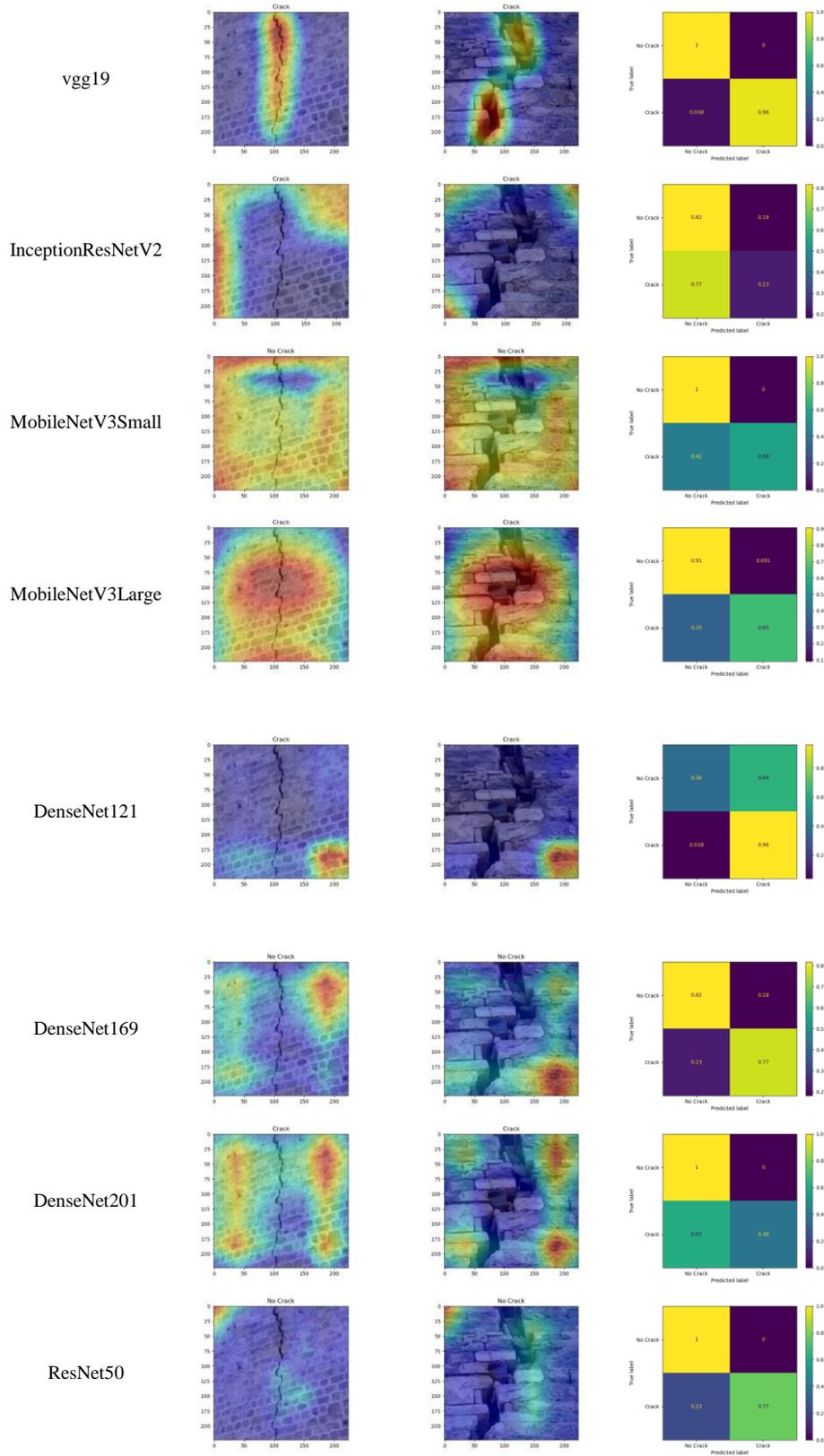

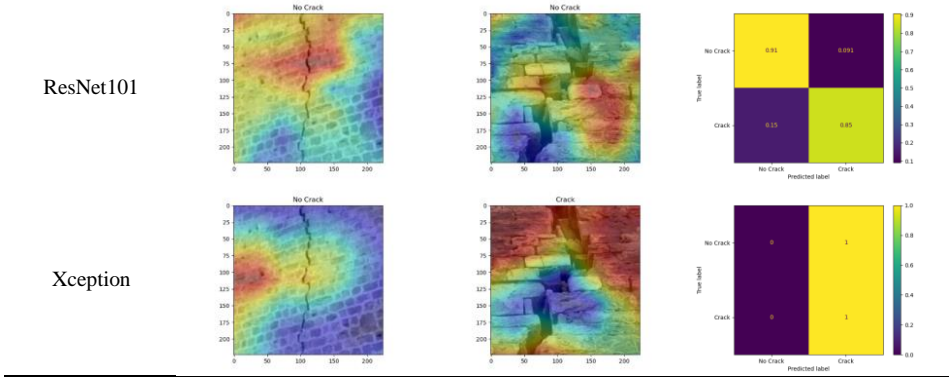

Fig. A.12: Evaluation results on the Rest test sites of the data using the pretrained models on ImageNet, after training only the additional Global Average Pooling (2D) and dense layers on the internet data

| model | eval_img_1 | eval_img_2 | cm |
|---|---|---|---|

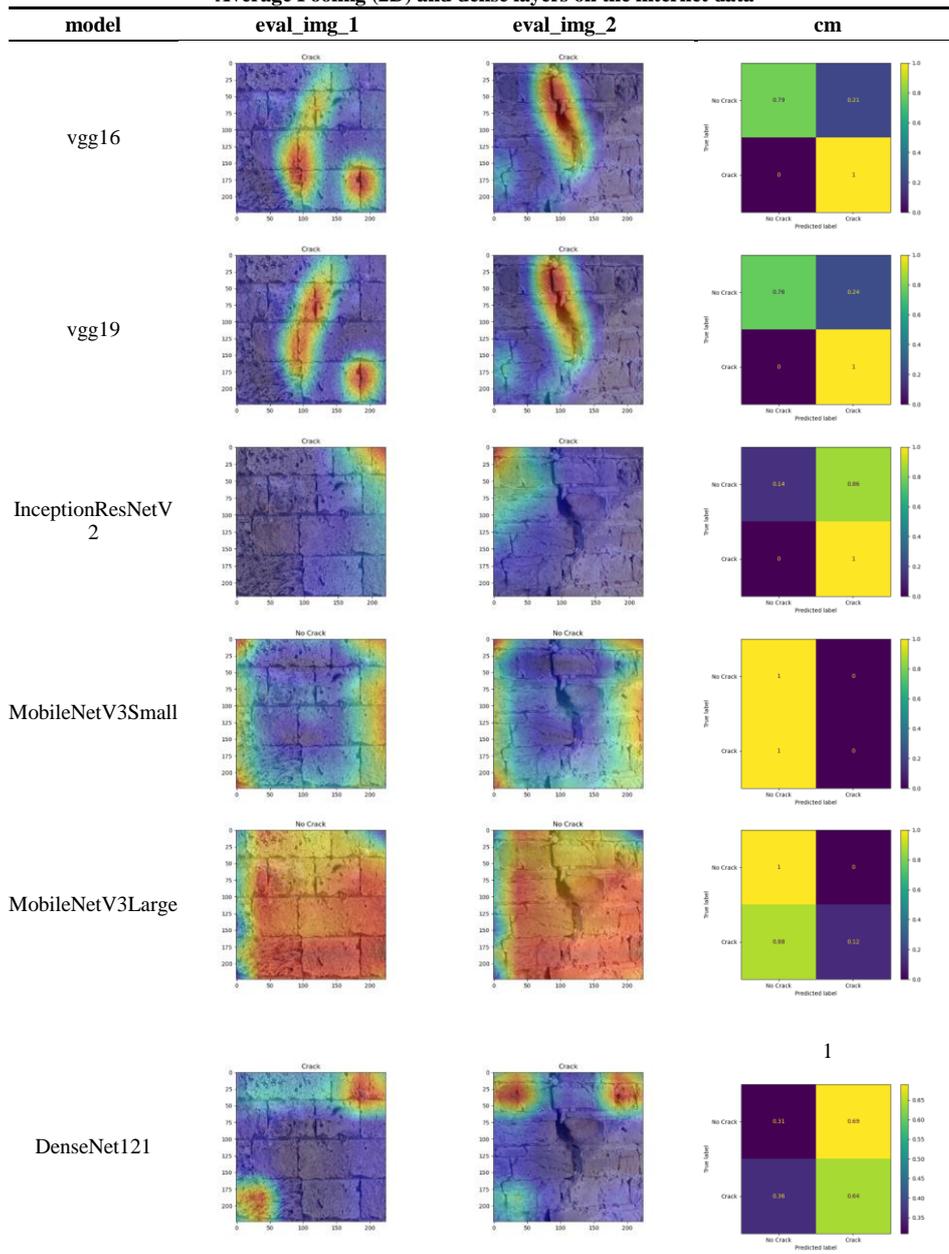

DenseNet169 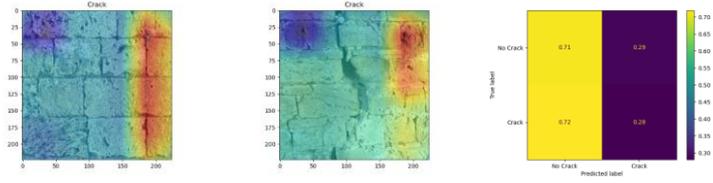

DenseNet201 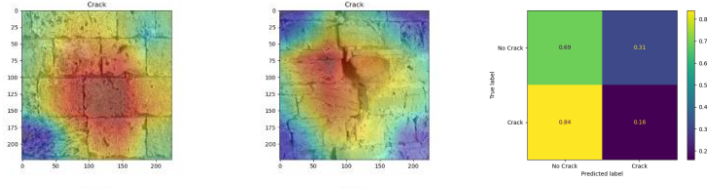

ResNet50 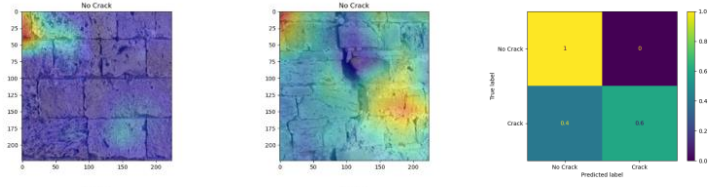

ResNet101 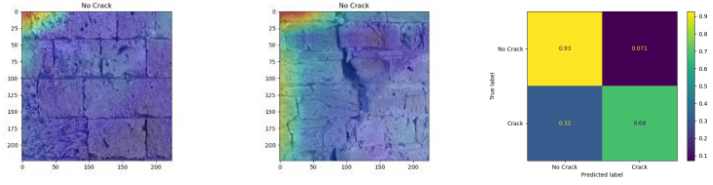

Xception 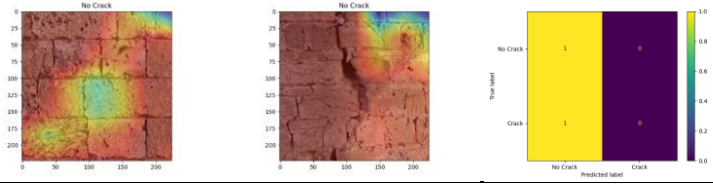